\documentclass[journal,onecolumn]{IEEEtran}

\usepackage{graphics} 
\usepackage{epsfig} 
\usepackage{mathptmx} 
\usepackage{times} 
\usepackage{amsmath} 
\usepackage{amssymb}  
\usepackage{makecell}
\usepackage{blindtext}
\usepackage{url}
\usepackage{xspace} 
\usepackage{enumitem}
\usepackage[justification=centering]{caption}
\usepackage{xcolor}
\usepackage{multirow}
\usepackage{graphbox}

\usepackage[utf8]{inputenc} 
\usepackage[T1]{fontenc}    
\usepackage{hyperref}       
\usepackage{url}            
\usepackage{booktabs}       
\usepackage{amsfonts}       
\usepackage{nicefrac}       
\usepackage{microtype}      
\usepackage{graphicx}           
\usepackage{subfigure}
\usepackage{tikz}
\usepackage{url}
\usepackage{amsmath} 
\usepackage{bm}      
\usepackage[numbers]{natbib}
\usepackage{array}
\usepackage{arydshln}
\usepackage{float}

\title{Knowledge-Guided Data-Centric AI in Healthcare: \\
Progress, Shortcomings, and Future Directions}
\date{}
\author{
  {\bf Edward Y. Chang} \\
  Stanford University \\
  echang@cs.stanford.edu \\
  Fellow of ACM \& IEEE
}

\newcommand{\ignore}[1]{}   

\newcommand{\LL}{\mathcal{L}}

\newcommand{\w}{\lambda}

\newcommand{\V}{\surd}

\newcommand{\y}{y}

\begin{document}
\maketitle
\begin{abstract}
The success of deep learning is largely due to the availability of large amounts of training data that cover a wide range of examples of a particular concept or meaning. In the field of medicine, having a diverse set of training data on a particular disease can lead to the development of a model that is able to accurately predict the disease. However, despite the potential benefits, there have not been significant advances in image-based diagnosis due to a lack of high-quality annotated data. This article highlights the importance of using a data-centric approach to improve the quality of data representations, particularly in cases where the available data is limited. To address this "small-data" issue, we discuss four methods for generating and aggregating training data: data augmentation, transfer learning, federated learning, and GANs (generative adversarial networks). We also propose the use of knowledge-guided GANs to incorporate domain knowledge in the training data generation process. With the recent progress in large pre-trained language models, we believe it is possible to acquire high-quality knowledge that can be used to improve the effectiveness of knowledge-guided generative methods.
\end{abstract}




\section{Introduction}
\label{dmd:introduction}


Applying artificial intelligence (AI) techniques to improve various aspects of healthcare, such as disease diagnosis and treatment recommendations, has made significant progress over the past half-century, but there are still several challenges to be overcome. This article presents the current state of progress in this field, identifies key challenges, and discusses promising directions for future research.

The term ``AI'' regained popularity after AlphaGo's success in 2015, but in the scientific domain, AI is simply a set of machine learning or statistical learning algorithms. According to a survey paper by W. B. Schwartz \cite{AIMedicine1987}, pattern recognition techniques (such as Boolean algebra and naive Bayesian models) were first used in healthcare in the 1960s to learn the weights of a set of decision factors for classification and disease prediction. A notable example was Stanford's Dentral system \cite{DENTRAL-1988}, which helped chemists identify unknown organic molecules. In the 1970s and 1980s, the machine learning community focused on using AI to solve clinical problems. Pathophysiological knowledge and reasoning were incorporated into the development of rule-based expert systems. The main approach was to match a patient's symptoms with stored profiles of findings for each disease. MYCIN [20] is a representative system for identifying bacteria causing severe infections and recommending antibiotics. Another system, INTERNIST-I \cite{MYCIN1976}, used a multiple hypotheses strategy to consider multiple disease candidates without prematurely converging on a single prediction. 
The rule-based approach to AI in healthcare was not successful due to a lack of comprehensive data and insufficient accuracy. These limitations prevented the deployment of any rule-based systems in the healthcare industry during this time period. Additionally, there was resistance from healthcare professionals to the adoption of these systems due to both procedural and legal concerns.

From 1990s to 2010s, the focus of AI in healthcare shifted to developing tools and algorithms to improve performance. These efforts were driven by two key insights: first, that AI systems in healthcare must be able to handle imperfect or missing data, and second, that small or insufficient training data is often the norm in healthcare and must be directly addressed. To address these challenges, researchers developed and improved upon techniques such as fuzzy set theory, Bayesian networks, clustering, and support vector machines (SVMs). Fuzzy set theory, for example, can handle data uncertainty and missing data, while Bayesian networks can consider correlations between parameters to reduce dependencies. SVMs, which use the kernel trick, can significantly reduce the amount of required training data by classifying data instances in the similarity space. Both SVMs and data clustering techniques, such as K-means and manifold learning, can reduce computation dimensionality and improve prediction accuracy with less training data.

Since 2010, the {\em data-centric} approach, which involves using large volumes of data to learn data representations, has become more widely accepted in the field of artificial intelligence~\cite{chang2011foundationsch2,krizhevsky2012imagenet}.  This approach involves using large amounts of data to learn data representations, and is exemplified by algorithms such as deep learning \cite{hinton2007learning,hinton2006fast} and transformers~\cite{NIPS2017_3f5ee243}. This approach differs from the model-centric approach, which relies on human-designed computer models or algorithms to extract features from images or documents. Unlike the model-centric approach, the data-centric approach involves learning features/representations from data, and more data can improve the quality of these representations. Additionally, the internal parameters of data-centric algorithms like multi-layer perceptrons and convolutional neural networks(CNNs)~\cite{CNN6795724} can be changed or learned based on the structure found in large datasets. The features of an image in a data-centric pipeline like a CNN can be affected by other images, while the features of an image in a model-centric pipeline like (SIFT)~\cite{SIFT_IJCV04} are invariant. The greater the quantity and variety of data available, the better the representations that can be learned by a data-centric pipeline. When a learning algorithm has been trained on a sufficient number of instances of an object under various conditions, such as different postures and partial occlusion, the features learned from this data will be more comprehensive.

Despite advancements in algorithms and data quality, AI is still not widely deployed in hospitals. The reasons can be understood from three pieces of documents: the ``indictment'' of A. R., Feinstein \cite{hand1985artificial} made in 1977, the experience of Google's negative deployment experience published in April 2020 \cite{MITTechReview-Google-2020}, and the comments made by Andrew Ng at Stanford Healthcare's AI Future workshop in April 2021 \cite{HAI-feifei-andrew-2021}. Their comments can be summarized into three factors: 1) a lack of deep interdisciplinary collaboration between computer scientists and clinicians, 2) a lack of robustness and interpretability in AI models, and 3) insufficient training data for the data-centric approach to learn good representations.  
This article focuses on addressing the training data issue.  
The current data augmentation techniques and transfer learning methods are not able to systematically enhance diversity 
to cover all variants of a medical condition. This leads to AI models being unable to effectively handle out-of-distribution instances. One
well known problem is that AI models trained in one hospital often do not perform well when used 
in a different hospital with different hardware and software configurations and clinical practices. 

In the remainder of this article we first summarize 
our prior studies in training-data 
generation~\cite{chang2019kggan,2017BookChap1,Transfer-EMBC2015,EURECOM+5916} and aggregation in 
Section~\ref{sec-tdata}. 
In Section~\ref{sec-kggan}, we propose a knowledge-guided generative model that 
can generate valid patterns not seen in the training data.
Finally, we enumerate 
plausible future research directions in Section~\ref{sec-conc}.

\section{Training Data Generation and Aggregation}
\label{sec-tdata}

The representation learning approach, which focuses on using data to train models, requires large and diverse training sets that can account for all possible variations of a target concept, such as a illness, object or document. However, collecting such data can be challenging, especially in healthcare. For example, annotating chest X-ray images for FDA certification requires three certified specialists, who are often in high demand and well paid. The task of annotating a set of 180,000 chest X-ray images, which includes the 108,938 NIH lung cancer dataset, took more than three years and was not completed by the end of 2020. Moreover, it is not possible for specialists to guarantee that these images represent all possible variations of chest diseases. To overcome these challenges, AI researchers have explored various methods to increase the volume and diversity of training data, such as data augmentation, federated learning, transfer learning, and data synthesis at the using methods such as GANs. In the rest of this section, we will examine each of these methods as well as their advantages and disadvantages through analyzes and our previous empirical studies~\cite{KGGAN8695311,2017BookChap1,Transfer-EMBC2015} .

\subsection{Data Augmentation}

It was observed by B. Li and E. Chang in 2002 in an image similarity study~\cite{DPF1040021} that an image that has gone through transformations such as scaling, cropping, down-sampling, color enhancement, lighting changes, etc. may appear to be perceptually similar to the original image. However, a traditional similarity function such as the Minkowski family function (e.g., L$1$ and L$2$ functions) would quantify these transformed images to be dissimilar to the original.  
There are three approaches to address this problem.  The first approach is to devise a distance function to correctly quantify the distance between perceptually similar images/objects.  Our proposed {\em Dynamic Partial Function} (DPF)  \cite{DPF1040021} is one effective method based on psychological principles. The second approach is
to employ a model that is insensitive to some image transformations.  For instance, the graph neural network (GNN) model~\cite{GNN9046288} connects key features of an object into a graph, which is invariant to transformations such as rotation and translation. However, GNN cannot model all transformations. One could consider employing GNN (instead of using CNN) as the training algorithm to work with augmented data.
Data augmentation is the third method, which generates all possible variants of a semantic (such as a disease and an object) and add them to the 
training data.  

AlexNet is a pioneer work that uses data augmentation extensively. According to \cite{krizhevsky2012imagenet}, AlexNet employs two distinct forms of data augmentation, both of which can produce the transformed images from the original images with very little computation \cite{krizhevsky2012imagenet,li2003discovery}.
The first scheme of data augmentation includes a random cropping function and horizontal reflection function. Data augmentation can be applied to both the training and testing stages. For the training stage, AlexNet randomly extracted smaller image patches $(224 \times 224)$ and their horizontal reflections from the original images $(256 \times 256)$. The AlexNet model was trained on these extracted patches instead of the original images in the ImageNet dataset. 
The second scheme of data augmentation alters the intensities of the RGB channels in training images by using principal component analysis (PCA). This scheme is used to capture an important property of natural images: the invariance of object identity to changes in the intensity and color of the illumination. 

Data augmentation enjoys popularity because of its simplicity. However, its major shortcomings are that
the generated new training instances may not appear in
the real world and that some unseen patterns of
an object cannot be generated via transformations.
Suppose data augmentation increases the number of training instances by $K$ folds.
The increased computational cost can be $O(K^2)$.

\subsection{Federated Learning}

Recently, federated learning attracts interests 
from the healthcare 
domain~\cite{Federated_MDPI} because of its 
advertised benefits of 
multi-source data integration,
privacy protection, and
and regulation compliance. 

Federated learning has shown 
limited success in rudimental computer
vision operations such as organ localization
and lesion segmentation~\cite{parekh2021crossdomain}.
Unfortunately, from both 
the architecture and practical perspectives,
federated learning faces tremendous challenges
for image-based disease diagnosis. We have
proposed a better architecture~\cite{Soteria9377624}, which
is summarized in the end of this subsection.

\begin{figure*}[ht!]
\begin{center}
\includegraphics[width=12.6cm,height=7.2cm]{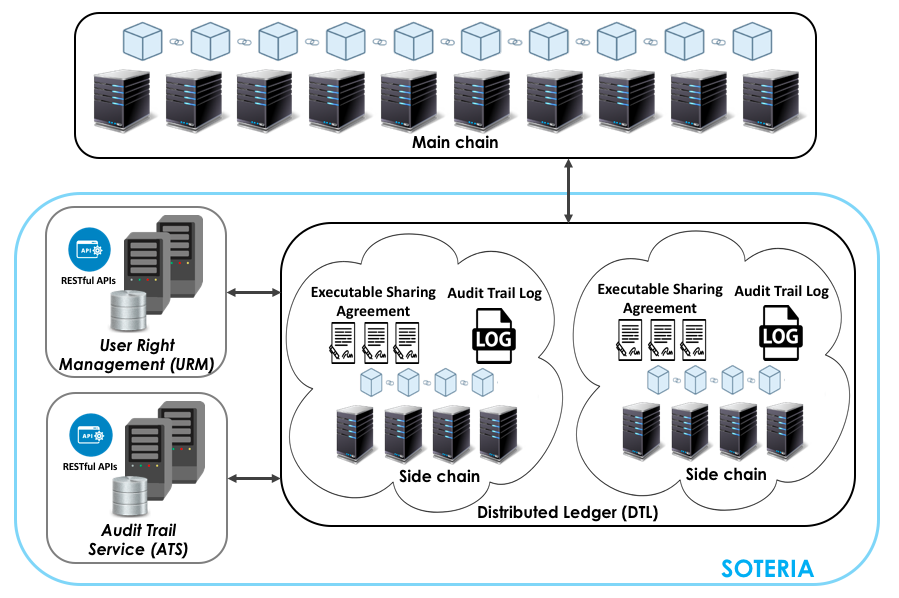}
\end{center}
\vspace{-.1in}
\caption{Soteria Components and its Three-layer Block Chains: Main, Side and Digital Agreement.}
\label{fig:soteria-arch}
\end{figure*}

Federal learning cannot overcome 
the following seven real-world multi-hospital
integration hurdles: 
\begin{itemize}
\item {\em Hardware diversity}. Heterogeneous hardware devices (e.g., MRI
machines) of different brands and models are used by 
different hospitals.
\item {\em Hardware parameters}.  Even with exactly the
same device, hardware parameter settings may be different,
which may result in different image size, resolution, and quality.
\item {\em Different clinical SOPs}. 
\item {\em Different DL models}. One hospital may have
a legacy system using GNN and another using CNN, etc..
\item {\em Different DL architectures}.  Since AlexNet, there has been double-digit DL architecture developed such as VGG, GoogleNet, ResNet, Inception, etc. It is virtually impossible
to ask all participants to use the same architecture.
\item {Different hyper-parameter values}. 
Even with the same architecture, the local optimal
hyper-parameter values may be different between sites.
\item {Different data format, quality, and annotation
practice}.
\end{itemize}

The fantasy to overcome all these seven 
issues requires draconian measures to
force all participating hospitals (domestic and
aboard) to use the same equipment, same hardware 
parameters, same photo-taking procedures, 
same image resolutions, 
same deep learning architectures,
same hyper-parameters, same image labeling conventions, etc.,
and output the parameters of the same latent layers for aggregation.

For text data, wherever the data are collected,
a word is a word.  But image analysis is 
very sensitive to the aforementioned variations.  Indeed,
Andrew Ng remakred at Stanford Healthcare's AI 
Future workshop in April 
2021 \cite{HAI-feifei-andrew-2021} that, 
a medical image model
trained at Stanford would simply fail to perform 
at a hospital down the street due to 
various variants.

How about privacy preservation?  Privacy regulations
such as GDPR~\cite{EUdataregulations2018}, CCPA~\cite{ccpabyCA20192020}, and HIPPA~\cite{HIPAA2017} 
require ``provable'' privacy 
preservation.  Federated learning is a close
system, and its privacy practice cannot be made
distributed, transparent, nor publicly provable. 
Our proposed Soteria architecture, shown
in Figure~\ref{fig:soteria-arch}, uses
a three layer blockchain-based ledger to support
publicly auditable privacy regulation compliance.  Moreover,
by placing digital contacts\footnote{A digital contract
is an agreement sign between the data owner and data 
consumer on the data-access consent with terms and
conditions. Terms and conditions can include 
access epoch and payment.
A digital contract is converted to a 
piece of SQL-like code to fetch data.} 
on the side chains, 
a data owner knows from the upper-layer blockchains
the status of his/her consent,
the access time to the data, and the payment for
each access, etc..  For details, please consult
the Soteria paper~\cite{Soteria9377624}.

\subsection{Transfer Representation Learning}
\label{sec-transfer}

The idea of transfer learning stems from
the fact that human beings can recognize 
a new object with just a small number
of examples. This few-shot learning capability
may come from our learned experience
in the past, which already well-tuned the parameters of
{\em our} brain.
It is also possible that
a pre-trained model in our cerebellum was given to us through heredity and prior learning~\cite{IEEEInfra-AGI-Consciousness, chang2023cocomo}. 
In machine learning, the common practice of transfer representation learning is to pre-train a CNN on a very large dataset (called the source domain) and then to use the pre-trained CNN either as an initialization or a fixed feature extractor for the task of interest  (called the target domain) \cite{CS231TL}. 

We use disease diagnosis as the target domain to illustrate the problems of and solutions to the challenges of small data training.
Specifically, we use otitis media (OM) and melanoma\footnote{In our award-winning XPRIZE Tricorder \cite{10.1145/3132635.3132637,tricorder_website} device (code name DeepQ), we effectively diagnose twelve conditions, and OM and melanoma are two of them.} as two example diseases.  
The training data available to us are 1) $1,195$ OM images collected by seven otolaryngologists at Cathay General Hospital\footnote{The dataset was used under a strict IRB process.  The dataset was deleted by April 2015 after our experiments had completed.}, Taiwan \cite{shie2014hybrid} and 2) $200$ melanoma images from PH$^2$ dataset \cite{mendoncca2013ph}.
The source domain from which representations are transferred to our two target diseases is  ImageNet~\cite{deng2009imagenet}.

\begin{figure}[t]
\vspace{-.1in}
  \centering
  \includegraphics[scale=.50]{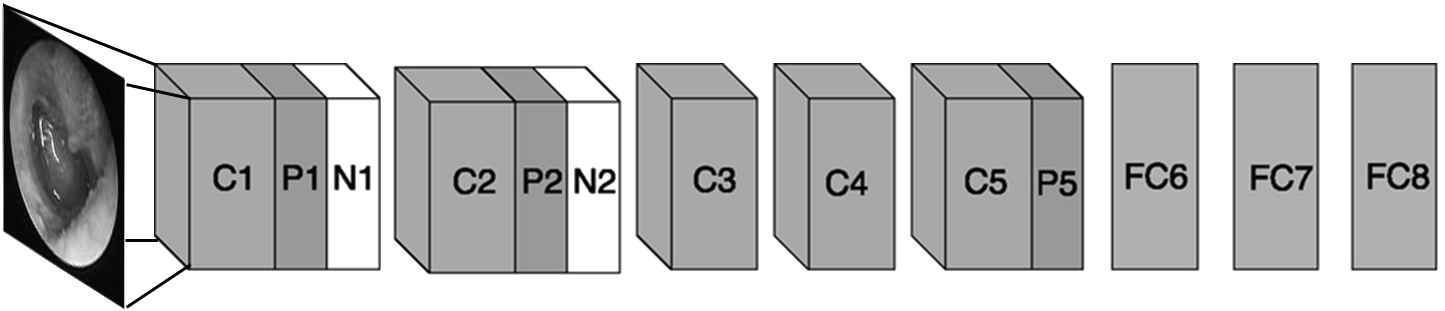}
  \caption{The Flowchart of Algorithm using OM Images as Example.}
  \label{f1}
\end{figure}

OM is any inflammation or infection of the middle ear, and treatment consumes significant medical resources each year \cite{american2004diagnosis}. Several symptoms such as redness, bulging, and tympanic membrane perforation may suggest an OM condition. Color, geometric, and texture descriptors may help in recognizing these symptoms. However, specifying these kinds of features involves a hand-crafted process and therefore requires domain expertise. Often times, human heuristics obtained from domain experts may not be able to capture the most discriminative characteristics, and hence the extracted features cannot achieve high detection accuracy. 
Similarly, melanoma, a deadly skin cancer, is diagnosed based on the widely-used dermoscopic ``ABCD'' rule~\cite{stolzABCD}, where A means asymmetry, B means border, C color, and D different structures. The precise identification of such visual cues relies on experienced dermatologists to articulate. Unfortunately, there are many congruent patterns shared by melanoma and nevus, with skin, hair, and wrinkles often preventing noise-free feature extraction.

Our transfer representation learning experiments consist of the following five steps:

\begin{enumerate}
\item Unsupervised codebook construction: We learned a codebook from ImageNet images, and this codebook construction is ``unsupervised'' with respect to OM and melanoma.
\item Encode OM and melanoma images using the codebook: Each image was encoded into a weighted combination of the pivots in the codebook. The weighting vector is the feature vector of the input image.
\item Supervised learning: Using the transfer-learned feature vectors, we then employed supervised learning to learn two classifiers from the $1,195$ labeled OM instances or $200$ labeled melanoma instances.
\item Feature fusion: We also combined some heuristic features of OM (published in \cite{shie2014hybrid}) and ABCD features of melanoma with features learned via transfer learning.
\item Fine tuning: We further fine-tuned the weights of the CNN using labeled data to improve classification accuracy.
\end{enumerate} 

As we will show in the remainder of this section, step four does not yield benefit, whereas the other steps are effective in improving diagnosis accuracy.  In other words, these two disease examples demonstrate that features modeled by domain experts or physicians (the model-centric approach) are ineffective. The data-centric approach of big data representation learning combined with small data adaptation is convincingly promising.


\subsection*{C.1 Method Specifications}

We started with unsupervised codebook construction. On the large ImageNet dataset, we learned the representation of these images using AlexNet~\cite{krizhevsky2012imagenet}.  AlexNet contains eight neural network layers. The first five are convolutional and the remaining three are fully-connected. Different hidden layers represent different levels of abstraction concepts. We utilized AlexNet in Caffe~\cite{jia2014caffe} as our foundation to build our encoder to capture generic visual features.

For each image input, we obtained a feature vector using the codebook. The information of the image moves from the input layer to the output layer through the inner hidden layers. Each layer is a weighted combination of the previous layer and stands for a feature representation of the input image. Since the computation is hierarchical, higher layers intuitively represent higher concepts. For images, the neurons from lower levels describe rudimentary perceptual elements like edges and corners, whereas the neurons from higher layers represent aspects of objects such as their parts and categories. To capture high-level abstractions, we extracted transfer-learned features of OM and melanoma images from the fifth, sixth and seventh layers, denoted as pool$5$ (P$5$), fc$6$ and fc$7$ in Fig.~\ref{f1} respectively.

Once we had transfer-learned feature vectors of the $1,195$ collected OM images and $200$ melanoma images, we performed supervised learning by training a support vector machine (SVM) classifier~\cite{chang2011libsvm}. We chose SVMs to be our model since it is an effective classifier widely used by prior works. Using the same SVM algorithm lets us perform comparisons with the other schemes solely based on feature representation. As usual, we scaled features to the same range and found parameters through cross validation. For fair comparisons with previous OM works, we selected the radial basis function (RBF) kernel.

\begin{figure}[t]
  \vspace{-.1in}
  \centering
  \includegraphics[scale=.42]{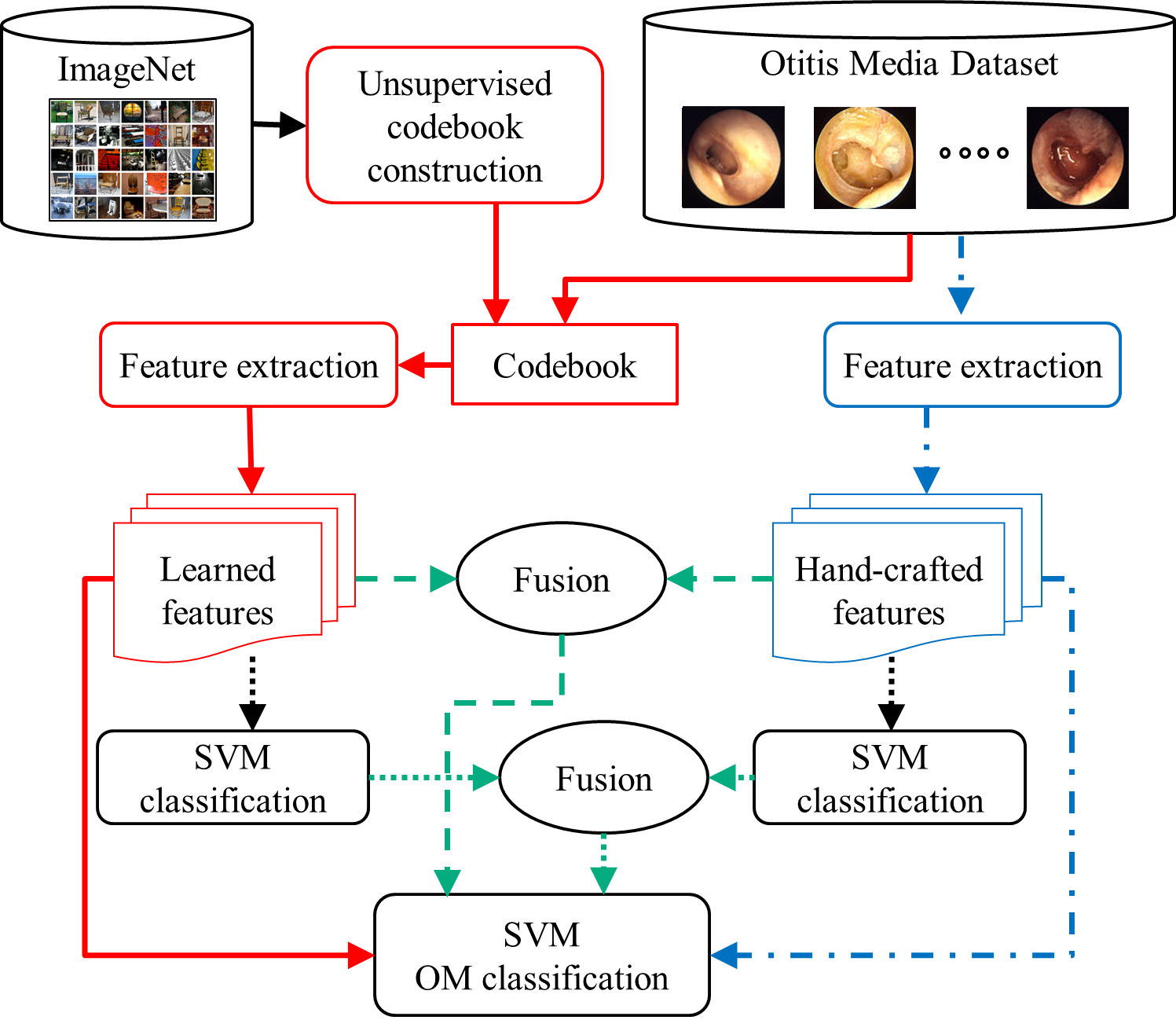}
  \caption{Four classification flows (OM photos are from~\cite{OM_website}).}
  \label{f2}
\end{figure}

To further improve classification accuracy, we experimented with two feature fusion schemes, which combine OM features hand-crafted by human heuristics (or model-centric)  in~\cite{shie2014hybrid} and our melanoma heuristic features with features learned from our codebook. In the first scheme, we combined transfer-learned and hand-crafted features to form fusion feature vectors. We then deployed the supervised learning on the fused feature vectors to train an SVM classifier. In the second scheme, we used the two-layer classifier fusion structure proposed in~\cite{shie2014hybrid}. In brief, in the first layer we trained different classifiers based on different feature sets separately. We then combined the outputs from the first layer to train the classifier in the second layer.

Fig.~\ref{f2} summarizes our transfer representation learning approaches using OM images as an example. The top of the figure depicts two feature-learning schemes: the transfer-learned scheme on the left-hand side and the hand-crafted scheme on the right. The solid lines depict how OM or melanoma features are extracted via the transfer-learned codebook, whereas the dashed lines represent the flow of hand-crafted feature extraction. The bottom half of the figure describes two fusion schemes. Whereas the dashed lines illustrate the feature fusion by concatenating two feature sets, the dotted lines show the second fusion scheme at the classifier level. At the bottom of the figure, the four classification flows yield their respective OM-prediction decisions. In order from left to right in the figure are 'transfer-learned features only', 'feature-level fusion', 'classifier-level fusion', and 'hand-crafted features only'.

\subsection*{C.2 Empirical Study and Discussion}
\label{om:experiment}

Two sets of experiments were conducted in our prior work~\cite{KGGAN8695311} 
to validate our idea. In this subsection, we first report OM classification performance by using our proposed transfer representation learning approach, followed by our melanoma classification performance. 
Then, we elaborate the correlations between images of ImageNet classes and images of disease classes by using a visualization tool to explain why transfer representation learning works.

For fine-tuning experiments, we performed a $10$-fold cross-validation for OM and a $5$-fold cross-validation for melanoma to train and test our models, so the test data are separated from the training dataset.  We applied data augmentation, including random flip, mirroring, and translation, to all the images.

For the setting of training hyperparameters and network architectures, we used mini-batch gradient descent with a batch size of $64$ examples, learning rate of $0.001$, momentum of $0.9$ and weight decay of $0.0005$. To fine-tune the AlexNet model, we replaced the fc$6$, fc$7$ and fc$8$ layers with three new layers initialized by using a Gaussian distribution with a mean of $0$ and a std of $0.01$. During the training process, the learning rates of those new layers were ten times greater than that of the other layers.

\medskip 
\noindent{\bf Results of Transfer Representation Learning for OM}
\medskip

Our $1,195$ OM image dataset encompasses almost all OM diagnostic categories: normal; AOM: hyperemic stage, suppurative stage, ear drum perforation, subacute resolution stage, bullous myringitis, barotrauma; OME: with effusion, resolution stage (retracted); COM: simple perforation, active infection.
Table~\ref{Table_OM} compares OM classification results for different feature representations. All experiments were conducted using $10$-fold SVM classification. The measures of results reflect the discrimination capability of the features.

The first two rows in Table~\ref{Table_OM} show the results of human-heuristic methods (hand-crafted), followed by our proposed transfer-learned approach. The eardrum segmentation, denoted as ‘seg’, identifies the eardrum by removing OM-irrelevant information such as ear canal and earwax from the OM images~\cite{shie2014hybrid}. The best accuracy achieved by using human-heuristic methods is around $80\%$. With segmentation (the first row), the accuracy improves $3\%$ over that without segmentation (the second row).

Rows three to eight show results of applying transfer representation learning. All results outperform the results shown in rows one and two, suggesting that the features learned from transfer learning are superior to that of human-crafted ones.

Interestingly, segmentation does not help improve accuracy for learning representation via transfer learning. This indicates that the transfer-learned feature set is not only more discriminative but also more robust. Among three transfer-learning layer choices (layer five (pool$5$), layer six (fc$6$) and layer seven (fc$7$)), fc$6$ yields slightly better prediction accuracy for OM. We believe that fc$6$ provides features that are more general or fundamental to transfer to a novel domain than pool$5$ and fc$7$ do. 

\begin{table}[t]
  \newcounter{magicrownumbers}
 \newcommand\rownumber{\stepcounter{magicrownumbers}\arabic{magicrownumbers}}
  \renewcommand\arraystretch{1.8}
  \caption{OM classification experimental results. (The best shown in \bf{bold}.)}
  \centering
\begin{tabular}{l|l|llll}
    \hline
    & Method     & Accuracy(std)   & Sensitivity  & Specificity  &  F\_1\\
    \hline
    \hline
    \rownumber & Heuristic w/ seg        & $80.11\%(18.8)$ & $83.33\%$  & $75.66\%$  &  $0.822$ \\ 
    \rownumber & Heuristic w/o seg       & $76.19\%(17.8)$ & $79.38\%$  & $71.74\%$  &  $0.790$  \\
    \hline
    \rownumber & Transfer w/ seg (pool$5$) & $87.86\%(3.62)$ & $89.72\%$  & $86.26\%$  &  $0.890$  \\
    \rownumber & Transfer w/o seg (pool$5$)& $88.37\%(3.41)$ & $89.16\%$  & $87.08\%$  &  $0.894$ \\
    \rownumber & Transfer w/ seg (fc$6$)   & $87.58\%(3.45)$ & $89.33\%$  & $85.04\%$  &  $0.887$  \\
    \rownumber & Transfer w/o seg (fc$6$)   & $88.50\%(3.45)$ & $89.63\%$  & $86.90\%$  &  $0.895$ \\
    \rownumber & Transfer w/ seg (fc$7$)    & $85.60\%(3.45)$ & $87.50\%$  & $82.70\%$  &  $0.869$ \\
    \rownumber & Transfer w/o seg (fc$7$)   & $86.90\%(3.45)$ & $88.50\%$  & $84.90\%$  &  $0.879$ \\
    \hline
    \rownumber & Feature fusion           & $89.22\%(1.94)$ & $90.08\%$  & $87.81\%$  &  $0.900$  \\   
    \rownumber & Classifier fusion        &  $89.87\%(4.43)$ & $89.54\%$  & {\bf{90.20\%}}   &  $0.898$  \\
    \rownumber & Fine-tune                & {\bf{90.96\%(0.65)}}  & {\bf{91.32\%}}  & {\bf{90.20\%}}  &  {\bf{0.917}} \\    
    \hline
    \bottomrule
  \end{tabular}
\label{Table_OM}
\end{table}

We also directly used the $1,195$ OM images to train a new AlexNet model.The resulting classification accuracy was only $71.8\%$, much lower than applying transfer representation learning. This
result confirms our hypothesis that even though CNN is a good model, with merely $1,195$ OM images (without the ImageNet images to facilitate feature learning), it cannot learn discriminative features. 

Two fusion methods, combining both hand-crafted and transfer learning features, achieved a slightly higher OM-prediction F1-score ($0.9$ over $0.895$) than using transfer-learned features only. This statistically insignificant improvement suggests that hand-crafted features do not provide much help.

Finally, we used OM data to fine-tune the AlexNet model, which achieves the best accuracy (see line $11$ in red). For fine-tuning, we replaced the original fc$6$, fc$7$ and fc$8$ layers with the new ones and used OM data to train the whole network without freezing any parameters. In this way, the leaned features can be refined and are thus more aligned to the targeted task.  This result attests that the ability to adapt representations to data is a critical characteristic that makes deep learning superior to the other learning algorithms.

\medskip
\noindent{\bf Results of Transfer Representation Learning for Melanoma}
\medskip
 
We performed experiments on the PH$^2$ dataset whose dermoscopic images were obtained at the Dermatology Service of Hospital Pedro Hispano (Matosinhos, Portugal) under the same conditions through the Tuebinger Mole Analyzer system using a magnification of $20$x. The assessment of each label was performed by an expert dermatologist.

Table~\ref{Melnoma_table} compares melanoma classification results for different feature representations. In Table~\ref{Melnoma_table}, all the experiments except for the last two were conducted by using $5$-fold SVM classification. The last experiment involved fine-tuning, which was implemented and evaluated by using Caffe. We also performed data augmentation to balance the PH$^2$ dataset ($160$ normal images and $40$ melanoma images) . 

Unlike OM, we found the low-level features to be more effective in classifying melanoma. Among three transfer-learning layer choices, pool$5$ yields a more robust prediction accuracy than the other layers do for melanoma. The deeper the layer is, the worse the accuracy becomes.  We believe that pool$5$ provides low-level features that are suitable for delineating texture patterns that depict characteristics of melanoma. 

Rows three and seven show that the accuracy of transferred features is as good as that of the ABCD rule method with expert segmentation. These results reflect that deep transferred features are robust to noise such as hair or artifacts.

We used melanoma data to fine-tune the AlexNet model and obtained the best accuracy $92.81\%$ since all network parameters are refined to fit the target task by employing back propagation. We also compared our result with the cutting-edge method, which reported $98\%$ sensitivity  and $90\%$ specificity on PH$^2$ ~\cite{barata2015melanoma}. Their method requires preprocessing such as manual lesion segmentation to obtain ``clean'' data. In contrast, we utilized raw images without conducting any heuristic-based preprocessing. Thus, deep transfer learning can identify features in an unsupervised way to achieve as good classification accuracy as those features identified by domain experts.

\begin{table}[t]
\caption{Melanoma classification experimental results. (The best shown in \bf{bold}.)}
  \newcounter{magicrownumberss}
 \newcommand\rownumber{\stepcounter{magicrownumberss}\arabic{magicrownumberss}}
  \renewcommand\arraystretch{1.8}
  \centering
  \begin{tabular}{l|l|lllll}
    \hline
    & Method     & Accuracy(std)   & Sens  & Spec  &  F\_1\\
    \hline
    \hline
    \rownumber & ABCD rule w/ auto seg        & $84.38\%(13.02)$ & $85.63\%$  & $83.13\%$  &  $0.8512$ \\ 
    \rownumber & ABCD w/ manual seg      & $89.06\%(9.87)$ & $90.63\%$  & $87.50\%$  &  $0.9052$  \\
    \rownumber & Transfer w/o seg (pool$5$)     & \textbf{$89.06\%(10.23)$} & $92.50\%$  & $85.63\%$  &  $0.9082$ \\
    \rownumber & Transfer w/o seg (fc$6$)       & $85.31\%(11.43)$ & $83.13\%$  & $87.50\%$  &  $0.8686$ \\
    \rownumber & Transfer w/o seg (fc$7$)       & $79.83\%(14.27)$ & $84.38\%$  & $74.38\%$  &  $0.8379$ \\
    \rownumber & Feature fusion               & $90.00\%(9.68)$ & $92.50\%$  & $87.50\%$  &  $0.9157$  \\   
    \rownumber & Fine-tune                    & \bf{92.81\%(4.69)} & \bf{95.00\%}  & \bf{90.63\%}  &  \bf{0.9300}  \\
 \hline   
 \bottomrule
  \end{tabular}
  \label{Melnoma_table}
\end{table}
 
\medskip
\noindent{\bf Qualitative Evaluation - Visualization}
\medskip

In order to investigate what kinds of features are transferred or borrowed from the ImageNet dataset, we utilized a visualization tool to perform qualitative evaluation. Specifically, we used an attribute selection method, SVMAttributeEval~\cite{guyon2002gene} with Ranker search, to identify the most important features for recognizing OM and melanoma. Second, we mapped these important features back to their respective codebook and used the visualization tool from ~\citet{yosinski2015understanding} to find the top ImageNet classes causing the high value of these features. By observing the common visual appearances shared by the images of the disease classes and the retrieved top ImageNet classes, we were able to infer the transferred features.

\begin{figure}[t]
  \centering
  \includegraphics[scale=.67]{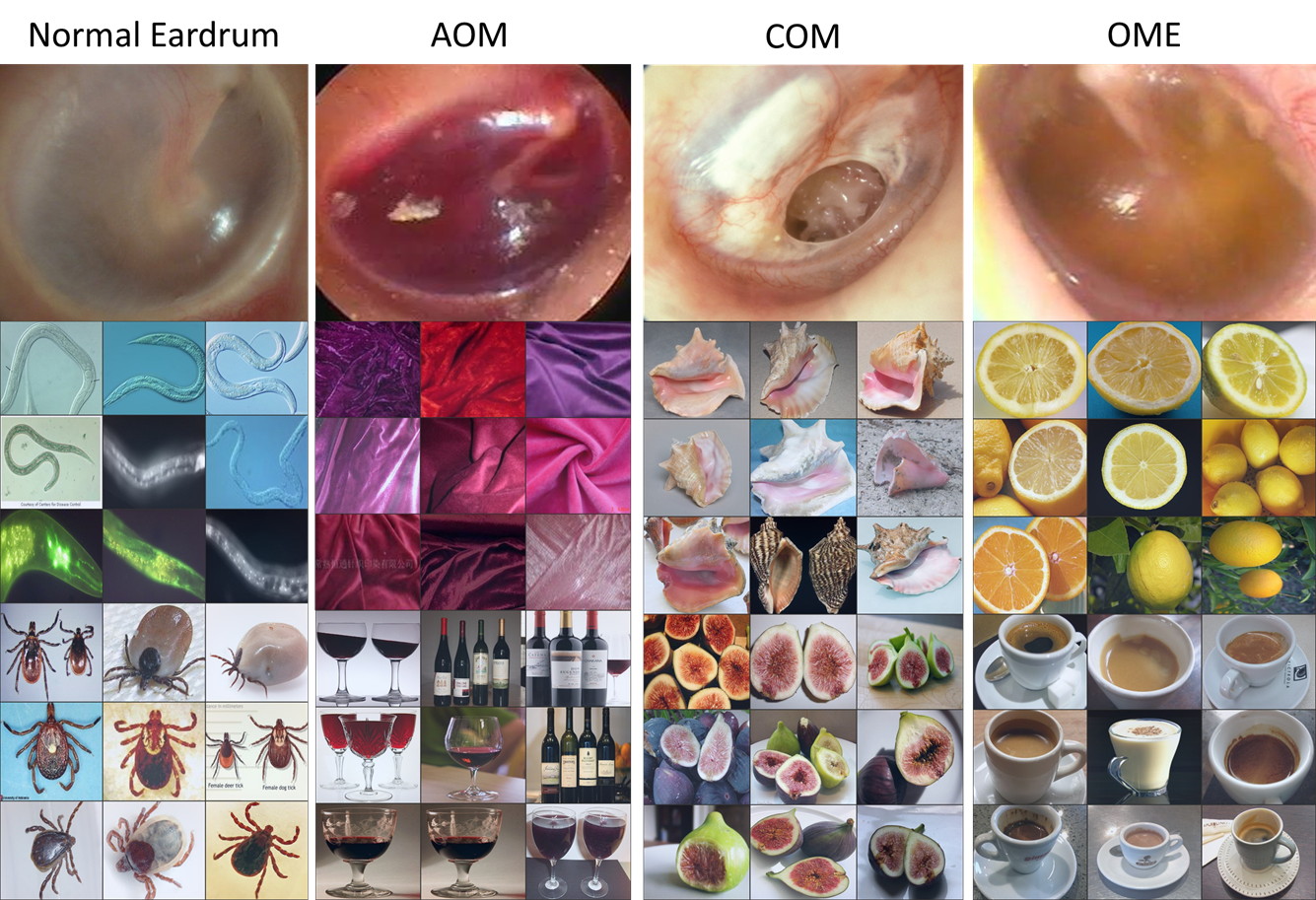}
  \caption{The visualization of helpful features from different classes corresponding to different OM symptoms (from left to right: Normal eardrum, AOM, COM, OME).}
  \label{f3}
\end{figure}

Fig.~\ref{f3} 
demonstrates the qualitative analyses of four different cases: the Normal eardrum, acute Otitis Media (AOM), Chronic Otitis Media (COM) and Otitis Media with Effusion (OME), which we will now proceed to explain in turn.  First, the normal eardrum, nematode and ticks are all similarly almost gray with a certain degree of transparency, features that are hard to capture with only hand-crafted methods. Second, AOM, purple-red cloth and red wine have red colors as an obvious common attribute. Third, COM and seashells are both commonly identified by a calcified eardrum. Fourth, OME, oranges, and coffee all seem to share similar colors. Here, transfer learning works to detect OM in an analogous fashion to how {\em explicit similes} are used in language to clarify meaning. The purpose of a simile is to provide information about one object by comparing it to something with which one is more familiar. For instance, if a doctor says that OM displays redness and certain textures, a patient may not be able to comprehend the doctor’s description exactly. However, if the doctor explains that OM presents with an appearance similar to that of a seashell, red wine, orange, or coffee colors, the patient is conceivably able to envision the appearance of OM at a much more precise level. At level fc$6$, transfer representation learning works like finding similes that can help explain OM using the representations learned in the source domain (ImageNet). 

The classification of melanoma contrasts sharply with the classification of OM. We can exploit distinct visual features to classify different OM classes. However, melanoma and benign nevi share very similar textures, as melanoma evolves from benign nevi. Moreover, melanoma often has atypical textures and presents in various colors. 

In the case of detecting melanoma versus benign nevus, effective representations of the diseases from higher-level visual characteristics cannot be found from the source domain. Instead, the most effective representations are only transferable at a lower-level of the CNN. We believe that if the source domain can add substantial images of texture-rich objects, the effect of explicit similes may be utilized at a higher level of the CNN.  For detailed analysis, readers can consult our work published in 2019~\cite{KGGAN8695311}.

\subsection*{C.3 Observations on Transfer Learning}
\label{om:conclusion}

The transfer learned features achieve accuracy $90.96\%$ ($91.32\%$ in sensitivity and $90.20\%$ in specificity) for OM and $92.81\%$ ($95.0\%$ in sensitivity and $90.63\%$ in specificity) for melanoma, achieving an improvement in disease-detection accuracy over the feature extraction instructed by domain experts. Moreover, our algorithms do not require manual data cleaning beforehand, and the preliminary diagnosis of OM and melanoma can be derived without aid from doctors. Therefore, automatic disease diagnosis systems, which hold the potential to help populations lacking in access to medical resources, are developmentally possible.


In summary, 
Our experiments on transfer learning provided three important insights on representation learning.

\begin{enumerate}
\item Low-level representations can be shared. Low-level perceptual features such as edges, corners, colors, and textures can be borrowed from some source domains where training data are abundant.  After all, low-level representations are similar despite different high-level semantics.

\item Middle-level representations can be correlated.  Analogous to explicit similes used in language, an object in the target domain can be ``represented'' or ``explained'' by some source domain features. In our OM visualization, we observed that a positive OM may display appearances similar to that of a seashell, red wine, oranges, or coffee colors --- features learned and transferred from the ImageNet source domain. 

\item Representations can adapt to a target domain. Even though, in the small data training situations, the amount of data is insufficient to learn effective representations by itself, given representations learned from some big-data source domains, the small data of the target domain can be used to align (e.g., re-weight) the representations learned from the source domains to adapt to the target domain.
However, predicting transferability from a source domain to a target domain is an open problem. Quantifying and predicting transferability remains to be a research problem to tackle. 
\end{enumerate}

\subsection{Generative Adversarial Networks (GANs)}
\label{sec-gans}

\begin{figure}[t]
\vspace{-.1in}
  \centering
  \includegraphics[scale=0.20]{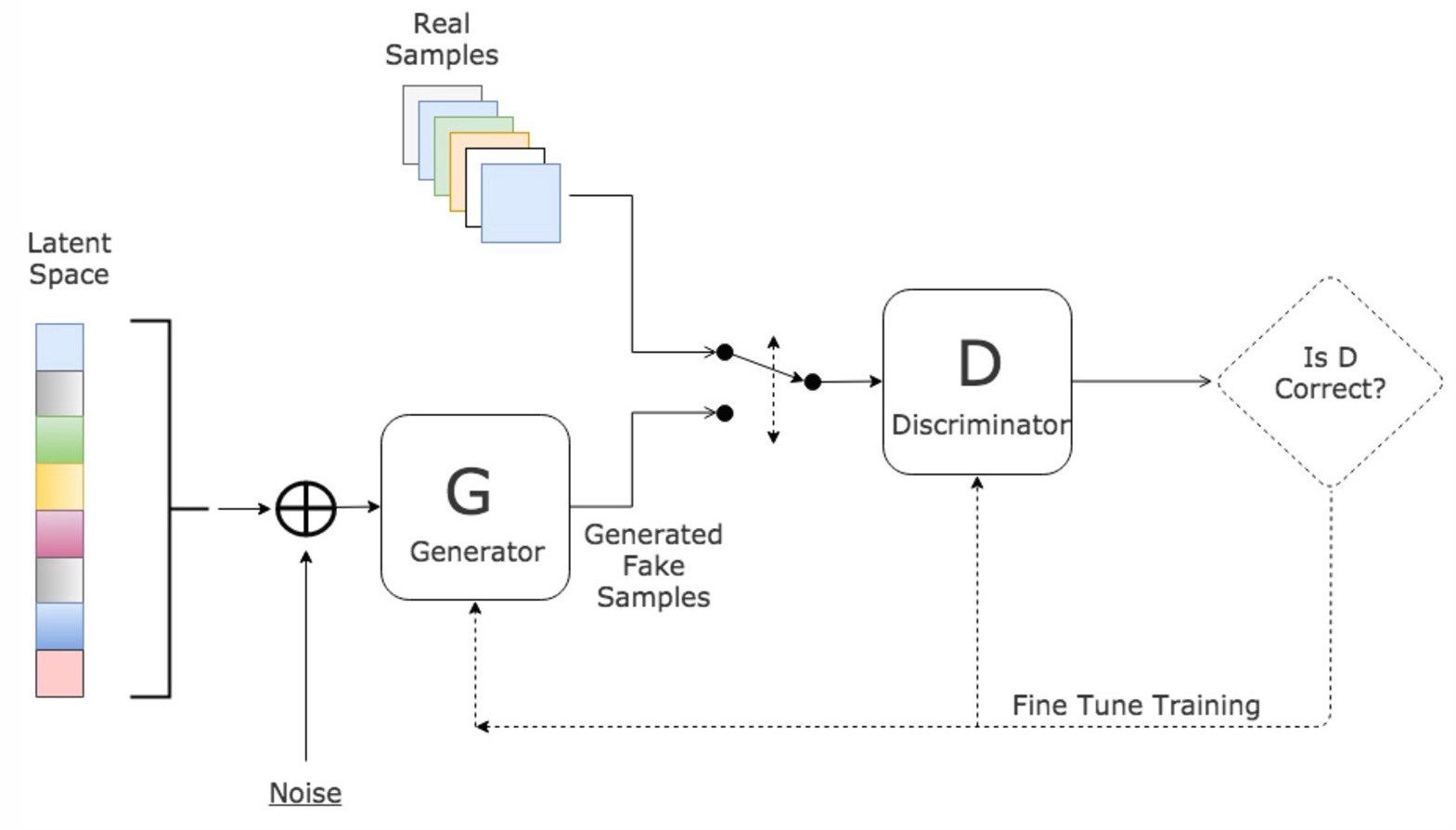}
  \caption{The Vanilla GANS by \cite{goodfellow2014generative}; figure credit: Hunter Heidenreich \cite{GAN-Drawing}.}
  \label{fig-gans}
\end{figure}

Generative adversarial networks (GANs) \cite{goodfellow2014generative} are a special type of neural network model where two networks are trained simultaneously.
Figure~\ref{fig-gans} depicts that the generator (denoted as $G$) focuses on producing fake images and the discriminator (denoted as $D$) centers on discriminating fake from real.
The goal is for the generator to produce fake images that can fool the discriminator to believe they are real.  
If an attempt fails, GANs use backpropagation to adjust network parameters. GANs have been used 
for transforming images to different styles~\cite{isola2017image}, or
changing facial expression of a person~\cite{RelGan9010872}.  
GANs have also been used to generate more training data.  

\subsection*{D.1 Method Specifications}

Since the introduction of the initial GAN model \cite{goodfellow2014generative}, there have been several variants depending on how the input, output, and error functions are modeled.
GANs can be primarily divided into four representative categories based on the input and output, and their applications are as follows:
\begin{itemize}
\item Conditional GAN (CGAN) \cite{mirza2014conditional}:
CGAN adds to GAN an additional input, y, on which the models can be conditioned.
Input y can be of any type, e.g., class labels.
Conditioning can be achieved by feeding y to both the generator $G(z|y)$ and the discriminator $D(x|y)$, where $x$ is a training instance and $z$ is random noise in latent space.
The benefit of conditioning on class labels is that it allows the generator to generate images of a particular class. (Application: text to image.)
\item Pixel-to-Pixel GAN (Pix2Pix) \cite{isola2017image}: Pix2Pix GAN is similar to CGAN. 
However, conditions are placed upon an image instead of a label.
The effect of such conditioning is that it allows the generator to map images of one style to another, e.g. mapping a photo to the painting style of an artist or mapping a sketch to a colored image. (Application: image to image translation, supervised.) 
\item Progressive-Growing GAN (PGGAN) \cite{karras2017progressive}:  
PGGAN grows both the generator and discriminator progressively; starting from low resolution, it adds new layers that model increasingly fine details as training progresses.
PGGAN can generate high-resolution images through progressive refinement. (Application: high-resolution image generation.)
\item Cycle GAN \cite{zhu2017unpaired}: 
Pix2Pix GAN requires paired training data to train.
Cycle GAN is an unsupervised approach for learning to translate an image from a source domain X to a target domain Y without training examples. 
The goal is to learn two mappings from $X$ to $Y$ (i.e., $G$) and from $Y$ to $X$ (i.e., $F$) such that the distributions $G(X)$ is indistinguishable from the distribution $Y$, and the distributions $F(Y)$ is indistinguishable from the distribution $X$, respectively. 
Cycle GAN introduces a cycle consistency loss to approximate $F(G(X))$ to X and also $G(F(Y))$ to Y. (Application: image to image translation, unsupervised.)
\end{itemize}


Though GANs have demonstrated interesting results, there are both micro and macro research issues that need to be addressed.

The micro issues are related to the formulation of the model's loss function to achieve good generalization.
But this generalization goal has been cast into doubt by the empirical study of \cite{arora2017generalization}, which concludes that training of GANs may not result in good generalization properties.
The GAN loss formulation is regarded as a saddle point optimization problem and training of the GAN is often accomplished by gradient-based methods \cite{goodfellow2014generative}.
$G$ and $D$ are trained alternatively so that they evolve together. 
However, there is no guarantee of balance between the training of G and D with the KL divergence. 
As a consequence, one network may inevitably be more powerful than the other, which in most cases, is D.
When D becomes too strong in comparison to G, the generated samples become too easy to differentiate from real ones.  
Another well known issue is that the two distributions are in high probability located in disjoint lower dimensional manifolds without overlaps.
The work of WGAN \cite{arjovsky2017wasserstein} addresses this issue by introducing the Wasserstein distance.  
However, WGAN still suffers from unstable training, slow convergence after weight clipping (when the clipping window is too large), and vanishing gradients (when the clipping window is too small).

The macro issue of GANs is: can GANs help generate large-volume and diversified training data to improve validation and testing accuracy?  
As stated in the introductory section, deep learning depends on the scale of training data to succeed, but most applications do not have ample training data.

Specifically in medical imaging, GANs have been mainly used in five areas: image reconstruction, synthesis, segmentation, registration, and classification, with hundreds of papers published since 2016 \cite{yi2018generative}.
A recent report \cite{ranschaert2018artificial} summarizes the state of applied AI in the field of radiology and conveys that promising results have been demonstrated, but the key challenge of data curation in collection, annotation, and management remains.
The work of \cite{frid2018gan} uses GANs to generate additional samples for liver lesion classification and claims that both the sensitivity and specificity are improved.
However, the total number of labeled images is merely $182$, which is too small a dataset to draw any convincing conclusions. 
The work \cite{salehinejad2018generalization} applies a similar idea to thoracic disease classification and achieves better performance.
The work uses human experts to remove noisy data, but fails to report how many noisy instances were removed and how much of the accuracy improvement was attributed to human intervention.
The paper also claims that additional data contributes in making training data of all classes balanced to mitigate the imbalanced training data issue.
Had the work demonstrated that generating additional data using GANs helps despite imbalanced distribution, the improved result would have been more convincing. 

Combining $3$D model simulation with GANs seems to be another plausible alternative to reaching the same goal of increasing training instances. 
The work of \cite{sixt2018rendergan} presents a framework that can generate a large amount of labeled data by combining a $3$D model with GANs. 
Another work \cite{shrivastava2017learning} combines a $3$D simulator (with labels) with unsupervised learning to learn a GAN model that can improve the realism of the simulating labeled data.
However, this combining scheme does not work for some tasks. For example, our AR platform Aristo \cite{zheng2017aristo} experimented with these methods and did not yield any accuracy improvements in its gesture recognition task.
Moreover, most medical conditions have lacked exact $3$D models so far, which makes the combining scheme difficult to apply.



\subsection*{D.2 Empirical Study and Discussion}

\begin{figure}[t]
\vspace{-.1in}
\begin{center}
\includegraphics[width=0.57\textwidth]{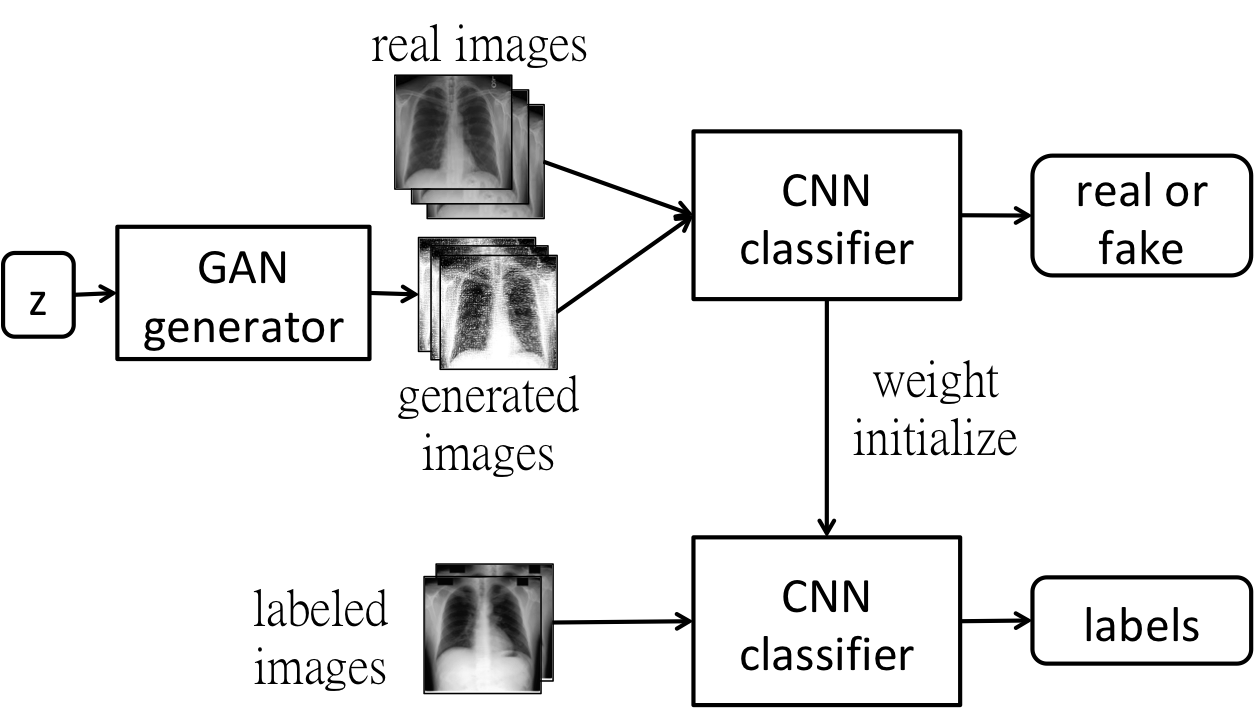}
\caption{Pre-Trained on Generated Images.}
\label{gan_pretraining}
\end{center}
\end{figure}

This section presents our experiments
performed in our prior work~\cite{KGGAN8695311} in generating training data using GANs to improve the accuracy of supervised learning.
Section~\ref{sec-transfer} shows that adding images unrelated to OM can improve classification accuracy due to representation transfer in the lower layers of the model and representation analogy in the middle layers of the model.
This leads us to the following questions: Can GANs produce useful labeled data to improve classification accuracy?
If so, which CNN layers can GANs strengthen to achieve the goal and how do GANs achieve this classification accuracy improvement?
Our experiments were designed to answer these questions.

\medskip
\noindent{\bf Experiment Setup }
\medskip

We used the NIH Chest X-ray 14 \cite{wang2017chestxray} dataset to conduct our experiments.
This dataset consists of $112,120$ labeled chest X-ray images, from over $30,000$ unique patients corresponding to 14 common thoracic disease types, including atelectasis, cardiomegaly, effusion, infiltration, mass, nodule, pneumonia, pneumothorax, consolidation, edema, emphysema, fibrosis, pleural thickening, and hernia.
The dataset is divided into training, validation, and testing sets, containing $78,468$, $11,219$. and $22,433$ images, respectively\footnote{We followed the dataset splits in \url{ https://github.com/zoogzog/chexnet/tree/master/dataset}}. 
Our experiments were designed to examine and compare four training methods:
\begin{enumerate}
    \item {Random initialization}: Model parameters were randomly initialized. 
    \item {\em Pre-trained by using ImageNet}: Similar to what
    we did with transfer learning in Section~\ref{sec-transfer}, the network was pre-trained by using ImageNet.
    \item {\em Pre-trained with additional data generated by unsupervised-GAN}: The method is shown in Figure~\ref{gan_pretraining}. First, the GAN generated the same number of fake images as we had real images. Second, the CNN classifier was trained to differentiate between real and fake images. Third, the weights were used to initialize the subsequent classification task. 
    \item {\em Trained with additional data generated 
    by supervised-GAN}: By adding the generated images, the size of the dataset was expanded to $2x$ and $5x$ (the size of original dataset is $x$).  In order to show whether GAN can produce labeled data to directly improve classification accuracy instead of indirectly, we changed the configuration of GAN in Method $3$ so that it could generate labeled images. 
\end{enumerate}
To establish a yardstick for these four methods, we first measured the ``golden'' results that supervised learning can attain using $100\%$ training and validation data.
We then dialed back the size of the training and validation data to be $50\%$, $20\%$, $10\%$, and then $5\%$.  
We used each of the four methods to either increase training data or pre-train the network.
We used PGGAN\footnote{We used a publicly available implementation of PGGAN via 
\url{https://github.com/tkarras/progressive_growing_of_gans}. This implementation has an auxiliary classifier \cite{odena2017conditional} and hence can generate images conditionally (for Method $4$) or unconditionally (for Method $3$).} as our GAN model to generate images with $1024 \times 1024$ pixel resolution.
For our CNN classifier, we employed DenseNet121 \cite{huang2017densely}, and used AUROC\footnote{We used a publicly available implementation of ChexNet \cite{rajpurkar2017chexnet} from \url{https://github.com/zoogzog/chexnet}, which contains a DenseNet121 classifier, and used its evaluation metric. The metric is deriving by first summing up all AUROCs from each of the 14 classes and then dividing the summation by 14.} as our evaluation metric.
Intuitively, our conjectures before seeing the results were as follows:

\begin{itemize}
    \item Method $1$ will perform the worst, since it does not
    receive any help to improve model parameters.
    \item Method $4$ will perform the best, since it produces
    more training instances for each target class.
    \item Method $3$ will outperform $2$ as the training data
    generated, though unlabeled, is more relevant to the target disease images than ImageNet is.
\end{itemize}

\medskip
\noindent{\bf Experiment Results}
\medskip

Table~\ref{experiment_table_gan} presents our experimental results. 
We report the AUROC of detecting $14$ thoracic disease types using each of the four different training methods.
These results are inconsistent with our conjectures:

\begin{table}[t]
 \newcommand\rownumber{\stepcounter{magicrownumbers}\arabic{magicrownumbers}}
  \renewcommand\arraystretch{1.8}
  \caption{Results of Four GANG Methods.}
  \centering
\begin{tabular}{l|lllll}
    \hline 
    \multicolumn{1}{c|}{ }  &
    \multicolumn{5}{c}{ Scale of Dataset} \\ \hline
    & 5\% & 10\% &20\% &50\% &100\% \\
    \hline \hline

    \makecell[l]{ Method 1 }
    &\makecell[l]{ 0.708 \\ (0.020) }  
    &\makecell[l]{ 0.757 \\ (0.003) }     
    &\makecell[l]{ 0.780 \\ (0.004) }   
    &\makecell[l]{ 0.807 \\ (0.002) }
    &\makecell[l]{ 0.829 \\ (0.000) }
    \\ \hline
    
    \makecell[l]{ Method 2 }
    &\makecell[l]{ 0.756 \\(0.006) }  
    &\makecell[l]{ 0.790 \\(0.002) }     
    &\makecell[l]{ 0.807 \\(0.005) }   
    &\makecell[l]{ 0.832 \\(0.001) }
    &\makecell[l]{ 0.843 \\(0.000) }
    \\ \hline
    
    \makecell[l]{ Method 3 }
    &\makecell[l]{ 0.726 \\ (0.002) }  
    &\makecell[l]{ 0.765 \\ (0.004) }    
    &\makecell[l]{ 0.789 \\ (0.001) }  
    &\makecell[l]{ 0.817 \\ (0.002) }
    &\makecell[l]{ 0.828 \\ (0.000) }
    \\ \hline

    \makecell[l]{ Method 4 (2x)  }
    &\makecell[l]{ 0.713 \\ (0.003) }   
    &\makecell[l]{ 0.724 \\ (0.004) }     
    &\makecell[l]{ 0.768 \\ (0.004) }  
    &\makecell[l]{ 0.809 \\ (0.001) } 
    &\makecell[l]{ 0.824 \\ (0.000) }
    \\ \hline
    
    \makecell[l]{ Method 4 (5x)  }
    &\makecell[l]{ 0.693 \\ (0.005) }   
    &\makecell[l]{ 0.727 \\ (0.002) }    
    &\makecell[l]{ 0.774 \\ (0.005) }   
    &\makecell[l]{ 0.798 \\ (0.005) }
    &\makecell[l]{ 0.813 \\ (0.000) }
    \\ \hline
    \bottomrule
  \end{tabular}
\label{experiment_table_gan}
\end{table}

\begin{itemize}
    \item Method $2$, which is equivalent to transfer learning, performs the best. No methods using GANs were able to outperform this method.
    \item Method $4$ performs the worst. In Method $4$, additional GAN-generated labeled images were used to perform training.  
    We believe that the labeled images generated using GANs were too noisy.  
    Therefore, when the generated images are increased ($5x$ vs. $2x$), the prediction accuracy is not always increased and sometimes even worse.
    This suggests that GANs do not produce helpful training instances and may in fact be counter-productive.
    \item Method $3$ does not outperform method $2$, even though ImageNet data used by method $2$ is entirely irrelevant to images of thoracic conditions.
    We believe that the additional images generated by GANs used for initializing network parameters are less useful because of their low volume and variety (diversity).
    After all, adding more low-quality similar images to an unlabeled pool cannot help the model learn novel features.
    Note that a recent keynote of I. Goodfellow
 \cite{GoodfellowAAAI2019} points out that GANs can successfully generate more unlabeled data (not labeled data) to improve MNIST classification accuracy. Table~\ref{experiment_table_gan} reflects the same conclusion that method $3$ outperforms
    method $1$, which uses randomly-initialized weights.  However, using GANs to generate
    unlabeled data may not be more productive than using ImageNet to pre-train the network.
\end{itemize}

\begin{figure}[t]
\centering
\resizebox{.88\textwidth}{!}{\begin{tabular}{ccc}
\includegraphics[width=0.36\textwidth]{img_real_1.png} &   
\includegraphics[width=0.36\textwidth]{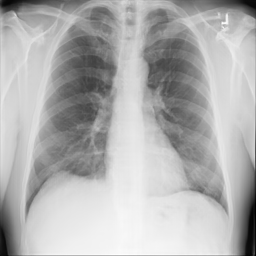} &
\includegraphics[width=0.36\textwidth]{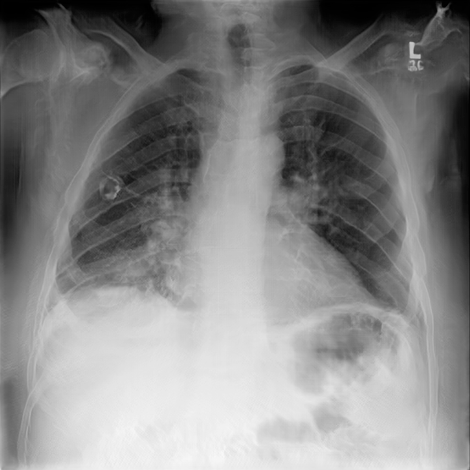} \\
\includegraphics[width=0.36\textwidth]{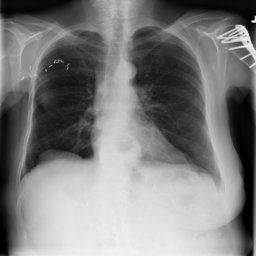} &   
\includegraphics[width=0.36\textwidth]{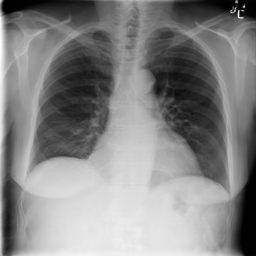} &
\includegraphics[width=0.36\textwidth]{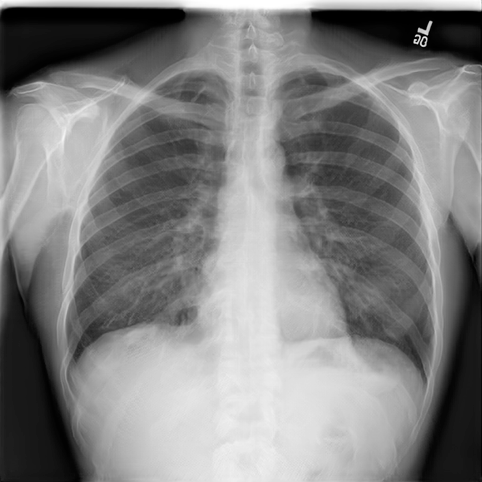} \\  
{Real} & {GAN unsupervised} & {GAN supervised} \\
\end{tabular}}
\caption{Real vs. GAN-Generated Images.  }
\vspace{-.1in}
\label{result_gan_images}
\end{figure}

Figure~\ref{result_gan_images} samples real and GAN-generated images.
The first column presents real images, the second column GAN-generated unsupervised, and the third 
GAN-generated supervised.
The GAN-generated images may successfully fool our colleagues with no medical knowledge.  However, as reported in \cite{salehinejad2018generalization}, the GAN-generated labeled chest X-ray images must be screened by a team of radiologists to remove erroneous data (with respect to diagnosis knowledge). Without domain knowledge, incorrectly labeled images may be introduced by GANs into the training pool, which would degrade classification accuracy.


\begin{figure}[ht]
\centering
\resizebox{.96\textwidth}{!}{\begin{tabular}{ccccc}
\includegraphics[width=0.18\textwidth]{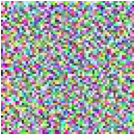} & 
\includegraphics[width=0.18\textwidth]{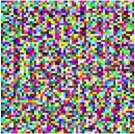} &
\includegraphics[width=0.18\textwidth]{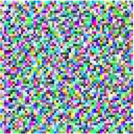} &
\includegraphics[width=0.18\textwidth]{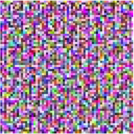} &
\includegraphics[width=0.18\textwidth]{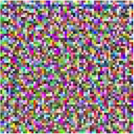}\\
\includegraphics[width=0.18\textwidth]{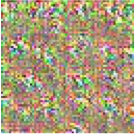} & 
\includegraphics[width=0.18\textwidth]{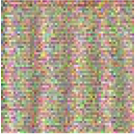} & 
\includegraphics[width=0.18\textwidth]{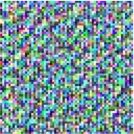} & 
\includegraphics[width=0.18\textwidth]{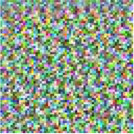} & 
\includegraphics[width=0.18\textwidth]{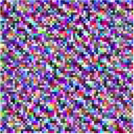}\\ 
\includegraphics[width=0.18\textwidth]{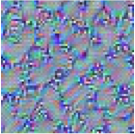} &
\includegraphics[width=0.18\textwidth]{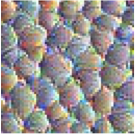} &
\includegraphics[width=0.18\textwidth]{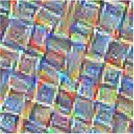} &
\includegraphics[width=0.18\textwidth]{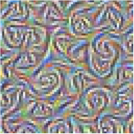} &
\includegraphics[width=0.18\textwidth]{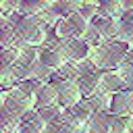}\\
\end{tabular}}
\caption{CNN layer visualization of the first denseblock of DenseNet121. The top row is random weight, the second row is pre-trained by unsupervised-GAN method, and the third row is pre-trained by ImageNet. } 
\vspace{-.1in}
\label{result_cnn_vis}
\end{figure}

In summary, the study of \cite{mahajan2018exploring} shows that pre-training with datasets that are multiple orders of magnitude larger than ImageNet can achieve higher performance than pre-training with only ImageNet on several image classification and object detection tasks.
This result further attests that volume and variety of data, even if unlabeled, helps improve accuracy. GANs may indeed achieve volume, but certainly cannot achieve variety.

To explain why using ImageNet can achieve better pre-training performance than that achieved when using GAN-generated images, we perform layer visualizations using the technique introduced in \cite{olah2017feature}.
Figure~\ref{result_cnn_vis} plots the output layer of the first dense-block of DenseNet. 
Row one shows five filters of untrained randomly initialized weights.
Row three shows five filters with more distinct features learned from the ImageNet pre-trained model.
The unsupervised-GAN method (row two) produces filters of similar quality to that of row one.
Qualitatively, unsupervised-GAN learns similar features akin to how the random-initialization method does, and does not yield more promising classification accuracy.



\section{Fusing Knowledge with GANs}
\label{sec-kggan}

The desired outcome of GANs after training is that samples formed by $x_g$ approximate the real data distribution $pr(x)$. 
However, if the real data distribution is under-represented by the training data, the generated samples cannot explore beyond the training data.  
For instance, if the otitis media (OM) training data shown in 
Section~\ref{sec-transfer} consists of 
only one type of OM, say AOM, GANs cannot generate the other two types of OM, COM and OME.
As another example, if a set of training data consists of a large number of red roses, and the aim of GANs is to generate entire categories of different colored roses, there would be no {\em knowledge} or {\em hint} for $G$ or $D$ to respectively achieve and tolerate diversity in color.
In other word, the discriminator $D$ would reject any roses that are not red and $G$ would not be encouraged to expand beyond generating red colored roses. The nature of GANs treats exploration beyond the paradigm of the seen or known to be erroneous.

If we would like GANs to generate diversified samples to improve supervised learning, the new approach must address two issues:

\begin{itemize}[leftmargin=.1in]
    \item Guiding the generator to explore diversity {\em productively}.
    \item Allowing the discriminator to tolerate diversity {\em reasonably}.
\end{itemize}

The adverbs {\em productively} {\em reasonably}, convey exploration (beyond exploitation) with guidance (via rules) and guardrails (via rewards, positive and negative).
In the case of playing games, rules and rewards are clear.
In the case of generating roses beyond red colors or generating types of flowers beyond roses, guidance and guardrails are difficult to articulate.
Supposing computer vision techniques can precisely segment petals of roses in an image, what colors can the generator $G$ use to replace red petals?  
For example, black roses do not exist, so this color would be deemed unreasonable and unproductive for generating realistic rose images. 
Exploration beyond training distribution should be permitted, 
but at the same time guided by knowledge.
How can knowledge be incorporated into training GANs? We enumerate two schemes.

\begin{enumerate}
\item {Incorporating a human in the loop}:
Placing a human in the loop instead of letting function $D$ make the decision can ensure $D$ is properly adjusted, due to human input.
The work of \cite{salehinejad2018generalization} discussed in Section~\ref{sec-gans} implements a GAN to generate labeled chest X-ray images and then asks a team of radiologists to remove mislabeled images.
We believe that merely removing ``bad'' images without productively generating new images with novel disease patterns may provide only limited help. 
\item {Encoding knowledge into GANs}:
We can convey to GANs about the information to be modeled via the knowledge layers/structures and/or via the knowledge graph/dictionary using natural language processing~\cite{chang2019kggan,KGGAN8695311}.
We elaborate this scheme in the remainder of this section.
\end{enumerate}

\subsection{Knowledge Acquisition Sources and Mechanisms}

Considering the structure of information may improve the effectiveness of GANs.
For instance, differentiating two types of strokes, ischemic and hemorrhagic, in order to provide proper treatment is critical for patient recovery.
Ischemic stroke, which accounts for $87$ percent of all stroke cases, occurs as a result of an obstruction within a blood vessel supplying blood to the brain. 
Hemorrhagic stroke occurs when a weakened blood vessel ruptures inside or on the surface of the brain. Two types of weakened blood vessels usually cause hemorrhagic stroke: aneurysms and arteriovenous malformations (AVMs). 

Without the above knowledge, GANs could generate data that flips the appearance of ischemic versus hemorrhagic strokes, which would blur the critical ability to differentiate between the two. 
Additionally, without knowledge of brain anatomy, GANs could generate obstructions and ruptures in clearly erroneous brain locations where no blood vessels are present.
With the knowledge that the symptoms largely occur within and on blood vessels, multi-layer GANs may be able to impose anatomical constrains through layering information. 

Let us use the rose example to explain two sources/mechanisms for knowledge acquisition.
Note that for each specific domain, e.g., nature and medicine, the knowledge sources can
be vastly different, but we hope that the acquisition mechanisms could be similar.  

\medskip
\noindent{\bf Wikipedia}
\medskip
    
The possible colors of roses can be obtained from the following 
Wikipedia text via natural language processing (NLP) parsing:
``Rose flowers have always been available in a number of colours and shades; they are also available in a number of colour mixes in one flower.
Breeders have been able to widen this range through all the options available with the range of pigments in the species. 
This gives us {\em yellow}, {\em orange}, {\em pink}, {\em red}, {\em white} and {\em many combinations of these colours}.
However, they lack the {\em blue pigment} that would give a true {\em purple} or blue colour and until the 21st century all true blue flowers were created using some form of dye.
Now, however, genetic modification is introducing the blue pigment.''

Once possible colors and their combinations have been extracted using NLP, we can enhance the idea of text-adaptive GANs \cite{nam2018text} to generate roses of these colors.  

\medskip
\noindent{\bf Large Pre-trained Models}
\medskip

Recent launches 
of ChatGPT \cite{ChatGPT_website} (on November 30${^{th}}$, 2022)
and DALL-E \cite{CALLE-openAI} by OpenAI demonstrate a pipeline
that one can generate images through prompting a large pre-trained language model.
For instance, GPT3~\cite{DBLP:journals/corr/abs-2005-14165} has 175-billion 
parameters and WuDao~\cite{WuDao_wiki, WuDao_ppt} has 1.75 trillions, 10 times of 
the GPT3's.
Though to-date, the iterative prompting and dialogue methods for acquiring
information are still primitive, users can already use DALL-E followed 
by prompting ChatGPT
to produce impressive results. In the end of this section, 
we present some examples in Figure~\ref{KGGAN-CALLE}, and discuss our
recent work in modeling 
consciousness~\cite{IEEEInfra-AGI-Consciousness,chang2023cocomo}
to make knowledge acquisition to support more 
effective and personalizable .

\subsection{Method Specifications}
\label{KGGAN-flowers}

\begin{figure*}
\begin{center}
	\includegraphics[width=.90\linewidth]{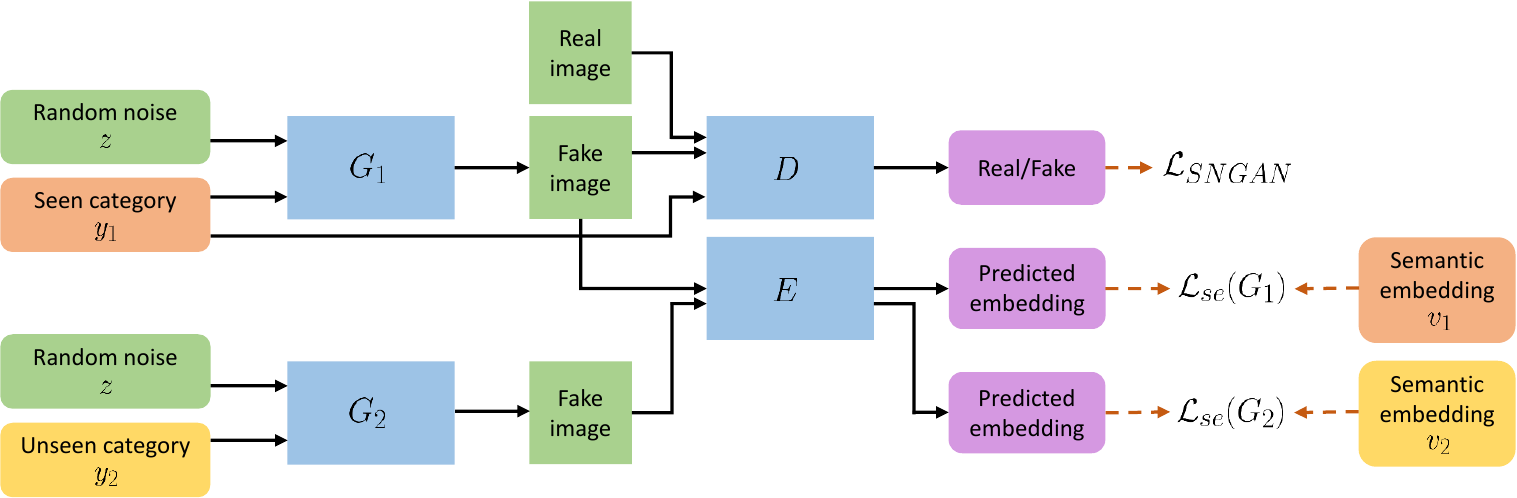}
	\caption{
	The schematic diagram of KG-GAN for unseen flower category generation.
	There are two generators $G_1$ and $G_2$, a discriminator $D$, and an embedding regression network $E$ as the constraint function $f$.
	We share all the weights between $G_1$ and $G_2$.
	By doing so, our method here can be treated as training a single generator with a category-dependent loss that seen and unseen categories correspond to optimizing two losses ($\LL_{SNGAN}$ and $\LL_{se}$) and a single loss ($\LL_{se}$), respectively, where $\LL_{se}$ is the semantic embedding loss.
	}
	\label{fig:flower_overview}
\end{center}
\end{figure*}

This section presents our proposed KG-GAN that incorporates domain knowledge into the GAN framework.
We consider a set of training data under-represented at the category level, i.e.,
all training samples belong to the set of seen categories, denoted as $Y_1$ (e.g.,
red category of roses),
while another set of unseen categories, denoted as $Y_2$ (e.g., any other
color categories), has no training samples.
Our goal is to learn categorical image generation for both $Y_1$ and $Y_2$.
To generate new data in $Y_1$, KG-GAN applies an existing GAN-based method to train a category-conditioned generator $G_1$ by minimizing GAN loss $\LL_{GAN}$ over $G_1$.
To generate unseen categories $Y_2$, KG-GAN trains another generator $G_2$ from the domain knowledge,
which is expressed by a constraint function $f$ that explicitly measures whether an image has the desired characteristics of a particular category.

KG-GAN consists of two parts:
(1) constructing the domain knowledge for the task at hand, and
(2) training two generators $G_1$ and $G_2$ that condition on available and unavailable categories, respectively.
KG-GAN shares the parameters between $G_1$ and $G_2$ to couple them together and to transfer knowledge learned from $G_1$ to $G_2$.
Based on the constraint function $f$, KG-GAN adds a 
\textit{knowledge loss}, denoted as $\LL_{K}$, to train $G_2$.
The general objective function of KG-GAN is written as 
$\min_{G_1, G_2} \LL_{GAN}(G_1) + \lambda\LL_{K}(G_2)$.



Given a flower dataset in which some categories are unseen,
our aim is using KG-GAN to generate unseen categories in addition to the seen categories.
Figure~\ref{fig:flower_overview} shows an overview of KG-GAN for unseen flower category generation.
Our generators take a random noise $z$ and a category variable $\y$ as inputs and generate an output image $x'$.
In particular, 
    $G_1: (z,\y_1) \mapsto x_1'$ 
and $G_2: (z,\y_2) \mapsto x_2'$, 
where $\y_1$ and $\y_2$ belong to the set of seen and unseen categories, respectively.

We leverage the domain knowledge that each category is characterized by a semantic embedding representation, 
which describes the semantic relationships among categories.
In other words, we assume that each category is associated with a semantic embedding vector $v$.
For example, we can acquire such feature representation from the textual descriptions of each category. 
(Figure~\ref{fig:dataset} shows example textual descriptions for four flowers.)
We use semantic embedding in two places:
one is for modifying the GAN architecture,
and the other is for defining the constraint function.
(Using the Oxford flowers dataset, we show how semantic embedding is done
in Section~\ref{sec:exp}.)


KG-GAN is developed upon SN-GAN~\cite{miyato2018cgans, miyato2018spectral}.
SN-GAN uses a projection-based discriminator $D$ and adopts spectral normalization for discriminator regularization.
The objective functions for training $G_1$ and $D$ use a hinge version of adversarial loss.
The category variable $\y_1$ in SN-GAN is a one-hot vector indicating which target category.
KG-GAN replaces the one-hot vector by the semantic embedding vector $v_1$.
By doing so, we directly encode the similarity relationships between categories into the GAN training.

The loss functions of the modified SN-GAN are defined as
\begin{equation}
    \begin{split}
        & \LL_{SNGAN}^G(G_1) = -\mathbb{E}_{z,v_1}[D(G_1(z,v_1),v_1)],~\text{and} \\
        & \LL_{SNGAN}^D(D) = \mathbb{E}_{x, v_1}[\max(0,1-D(x,v_1))] 
                         + \mathbb{E}_{z, v_1}[\max(0,1+D(G_1(z,v_1),v_1))].
    \end{split}
\end{equation}

\textbf{Semantic Embedding Loss.}
We define the constraint function $f$ as predicting the semantic embedding vector of the underlying category of an image.
To achieve that, we implement $f$ by training an embedding regression network $E$ from the training data.
Once trained, we fix its parameters and add it to the training of $G_1$ and $G_2$.
In particular, we propose a semantic embedding loss $\LL_{se}$ as the role of knowledge loss in KG-GAN.
This loss requires the predicted embedding of fake images to be close to the semantic embedding of target categories.
$\LL_{se}$ is written as
\begin{equation}
    \begin{split}
        \LL_{se}(G_i) & = \mathbb{E}_{z, v_i} || E(G_i(z, v_i)) - v_i ||^2, \text{where}~i \in \{1, 2\}.
    \end{split}
\end{equation}

\textbf{Total Loss.}
The total loss is a weighted combination of $\LL_{SNGAN}$ and $\LL_{se}$.
The loss functions for training $D$ and for training $G_1$ and $G_2$ are respectively defined as
\begin{equation}
    \begin{split}
        \LL^D & = \LL_{SNGAN}^D(D),~\text{and} \\
        \LL^G & = \LL_{SNGAN}^G(G_1) + \w_{se} (\LL_{se}(G_1) + \LL_{se}(G_2)).
    \end{split}
\end{equation}
\subsection{Empirical Study and Discussion} \label{sec:exp}

\begin{figure*}[t]
\begin{center}
	\includegraphics[width=.80\linewidth]{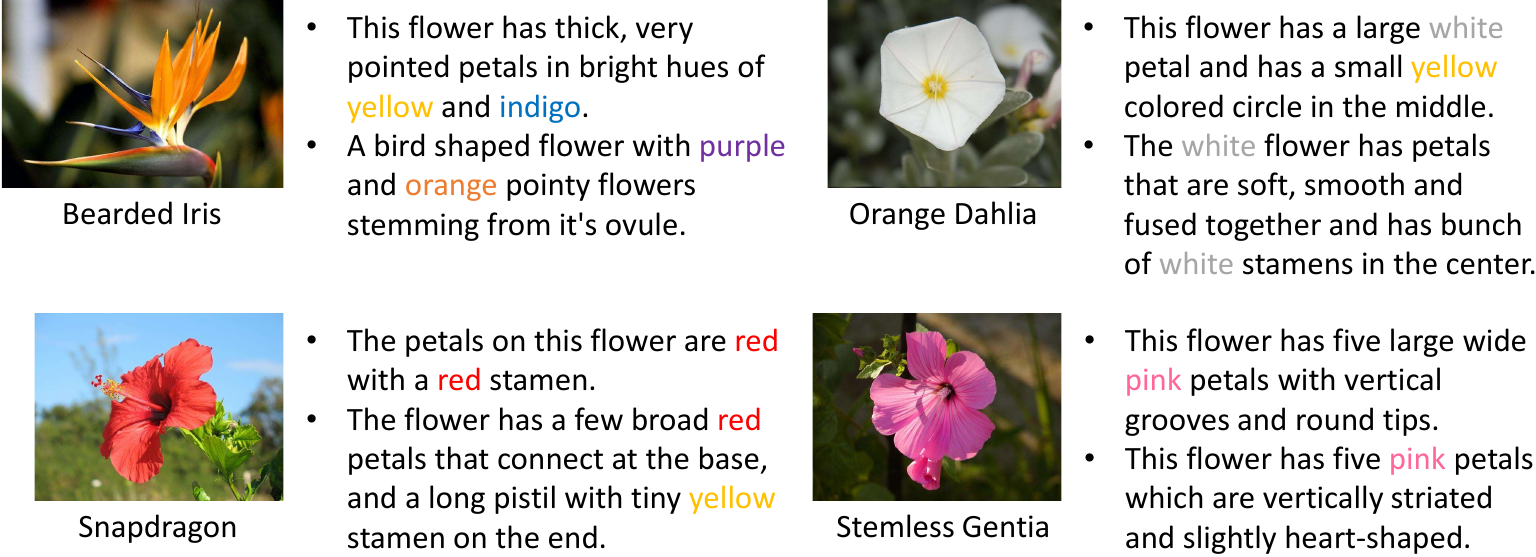}
	\caption{
	\textbf{Oxford flowers dataset}. Example images and their textual descriptions.
	}
	\label{fig:dataset}
\end{center}
\end{figure*}

\begin{figure*}[t]
\begin{center}
	\includegraphics[width=0.80\linewidth]{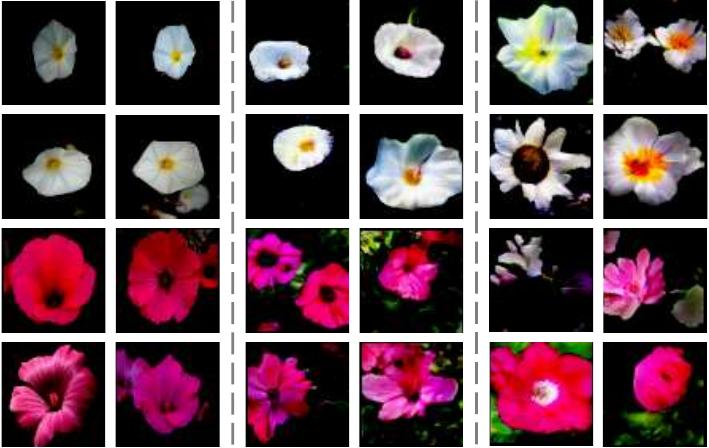}
    \caption{\textbf{Unseen flower category generation.} Qualitative comparison between real images and the generated images from KG-GAN.
    \textit{Left}: Real images.
    \textit{Middle}: Successful examples of KG-GAN.
    \textit{Right}: Unsuccessful examples of KG-GAN.
    The top two and the bottom two rows are Orange Dahlia and Stemless Gentian, respectively.
    }
	\label{fig:flower_results}
\end{center}
\end{figure*}

We use the Oxford flowers dataset~\cite{nilsback2008automated},
which contains $8{,}189$ flower images from $102$ categories (e.g., bluebell, daffodil, iris, and tulip).
Each image is annotated with $10$ textual descriptions.
Figure~\ref{fig:dataset} shows two representative descriptions for four flowers.
Following~\cite{reed2016learning}, we randomly split the images into $82$ seen and $20$ unseen categories.
To extract the semantic embedding vector of each category, 
we first extract sentence features from each textual description using the fastText library~\cite{bojanowski2017enriching},
which takes a sentence as input and outputs a $300$-dimensional real-valued vector in range $[0, 1]$. 
Then we average over the features within each category to obtain the per-category feature 
vector as the semantic embedding.
We resize the images to $64\times 64$ as the image size in our experiments.
For the SN-GAN part of the model, we use its default hyper-parameters and training configurations.
In particular, we train for $200$k iterations.
For the knowledge part, we use $\w_{se}=0.1$ in our experiments.

\textbf{Comparing Methods.}
We compare with SN-GAN trained on the full Oxford flowers dataset, 
which potentially represents a performance upper-bound of our method.
Besides, we additionally evaluate two ablations of KG-GAN:
(1) One-hot KG-GAN: $y$ is a one-hot vector that represents the target category.
(2) KG-GAN w/o $\LL_{se}$: our method without $\LL_{se}$.


\textbf{Results.} To evaluate the quality of the generated images, 
we compute the FID scores~\cite{heusel2017gans} in a per-category manner as in~\cite{miyato2018cgans}.
Then, we average over the FID scores of the set of the seen and the unseen categories, respectively.
Table~\ref{tab:part1_fid} shows the seen and the unseen FID scores.
We can see from the table that in terms of the category condition, 
semantic embedding gives better FID scores than one-hot representation.
Our full method achieves the best FID scores.
In Figure~\ref{fig:flower_results},
we show example results of two representative unseen categories.


\begin{table}[h]
    \small
    \caption{Per-category FID scores of SN-GAN and KG-GANs.}
    \label{tab:part1_fid}
    \centering
    \begin{tabular}{lcccccc}
    \toprule
    Method & Training data & Condition & $\LL_{se}$ & Seen FID & Unseen FID \\
    \midrule
    SN-GAN                & $Y_1\cup Y_2$ & One-hot   &            & 0.6922   & 0.6201 \\ 
    \midrule
    One-hot KG-GAN        & $Y_1$         & One-hot   & $\V$       & 0.7077   & 0.6286 \\ 
    KG-GAN w/o $\LL_{se}$ & $Y_1$         & Embedding &            & 0.1412   & 0.1408 \\ 
    KG-GAN                & $Y_1$         & Embedding & $\V$       & $\mathbf{0.1385}$ & $\mathbf{0.1386}$ \\ 
    \bottomrule
    \end{tabular}
\end{table}

\subsection{Observations on KG-GANs}

From Table~\ref{tab:part1_fid} we make two observations.
First, KG-GAN (conditioned on semantic embedding) performs better than One-hot KG-GAN.
This is because 
One-hot KG-GAN learns domain knowledge only from the knowledge constraint while KG-GAN additionally learns the similarity relationships between categories through the semantic embedding as the condition variable.
Second, when KG-GAN conditions on semantic embedding, 
KG-GAN with out $\LL_{se}$ still works.
This is because KG-GAN learns how to \textit{interpolate} 
among seen categories to generate unseen categories.
For example,
if an unseen category is close to a seen category in the semantic embedding space, 
then their images will be similar.

As we can see from Figure~\ref{fig:flower_results},
our model faithfully generates flowers with the right color, but does not perform as well
in shapes and structures.
The reasons are twofold.
First, colors can be more consistently articulated on a flower. Even if
some descriptors annotate a flower as red and while others annotate it as pink, 
we can obtain a relatively consistent color depiction over, say,
ten descriptions.  Shapes and structures do not enjoy 
as confined a vocabulary set as colors do.  In addition, the flowers in the same category may 
have various shapes and structures due to aging and camera angles.
Since each image has $10$ textual descriptions and each category has an average number of $80$ images, 
the semantic embedding vector of each category is obtained from taking an average 
over about $800$ fastText feature vectors.
This averaging operation preserves the color information quite well 
while blurring the other aspects.



\subsection{Knowledge Acquisition}

\begin{figure*}
\vspace{-.1in}
\begin{center}
	\includegraphics[width=.80\linewidth]{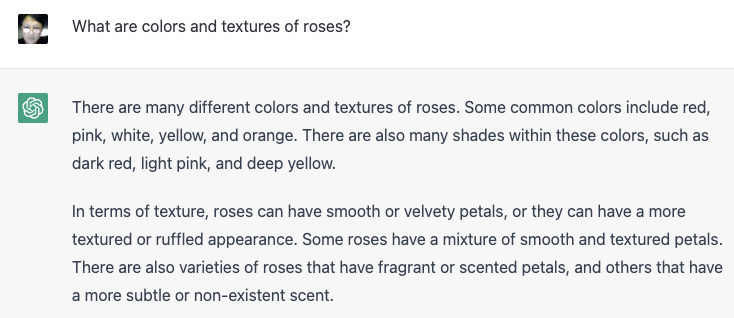}
	\caption{An Example ``Rose'' Query to ChatGPT.}
\label{fig:chatgpt-rose}
\end{center}
\vspace{-.1in}
\end{figure*}

\begin{figure}[h]
\centering
\resizebox{.88\textwidth}{!}{\begin{tabular}{ccc}
\includegraphics[width=0.50\textwidth]{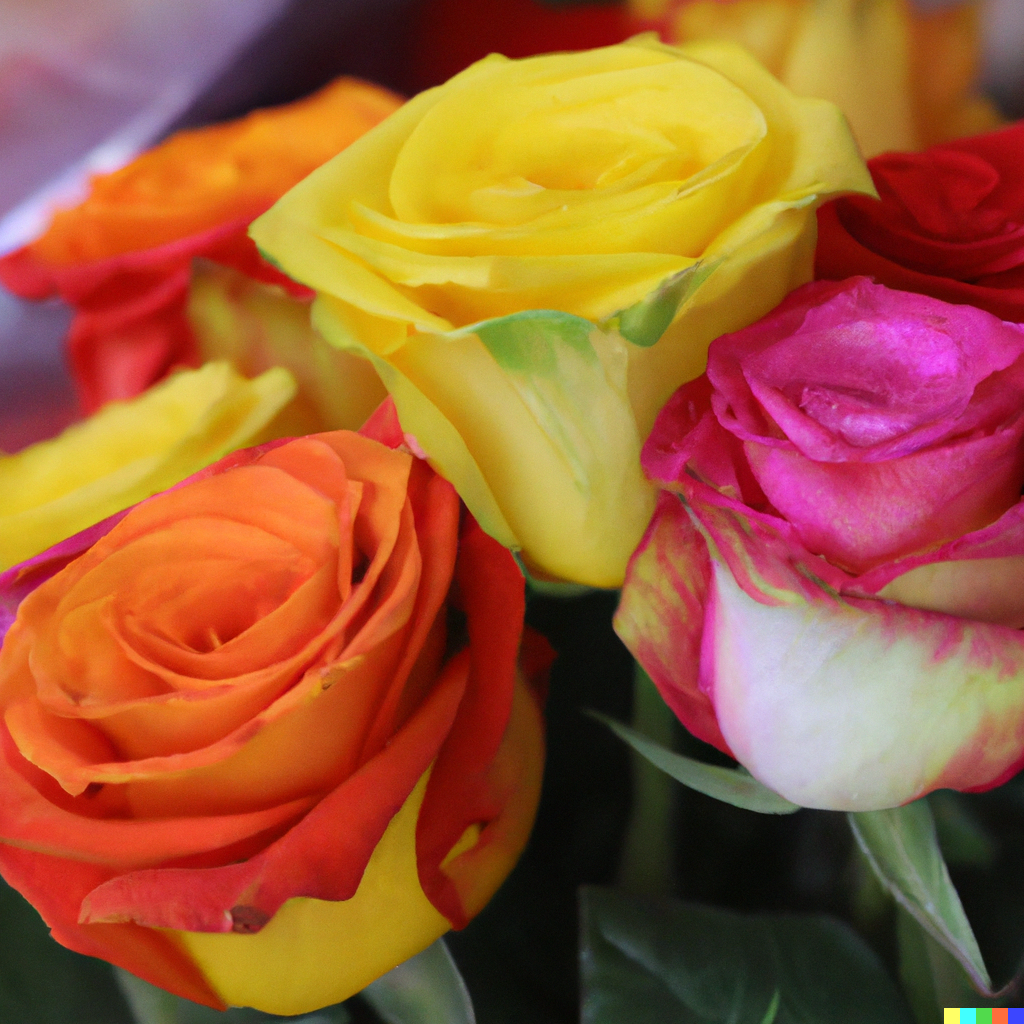} &   
\includegraphics[width=0.50\textwidth]{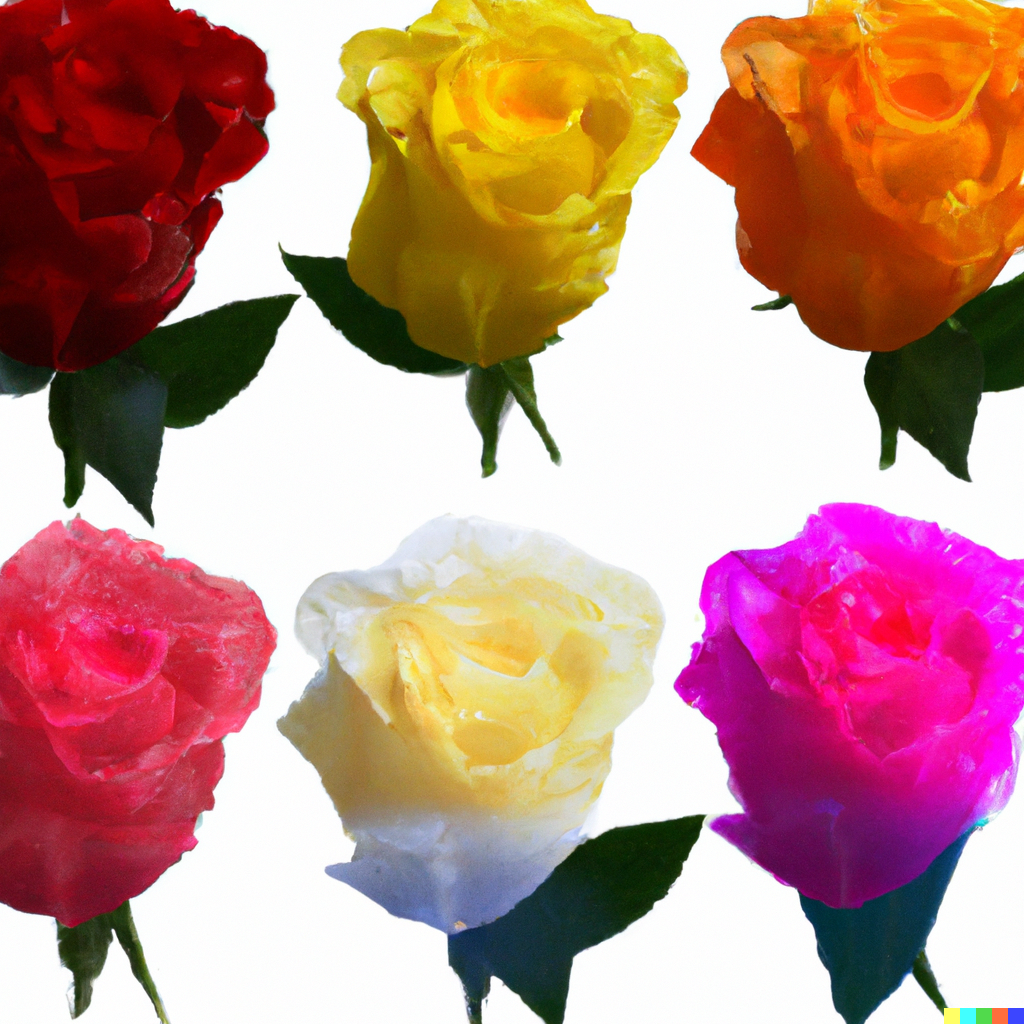} &
\includegraphics[width=0.50\textwidth]{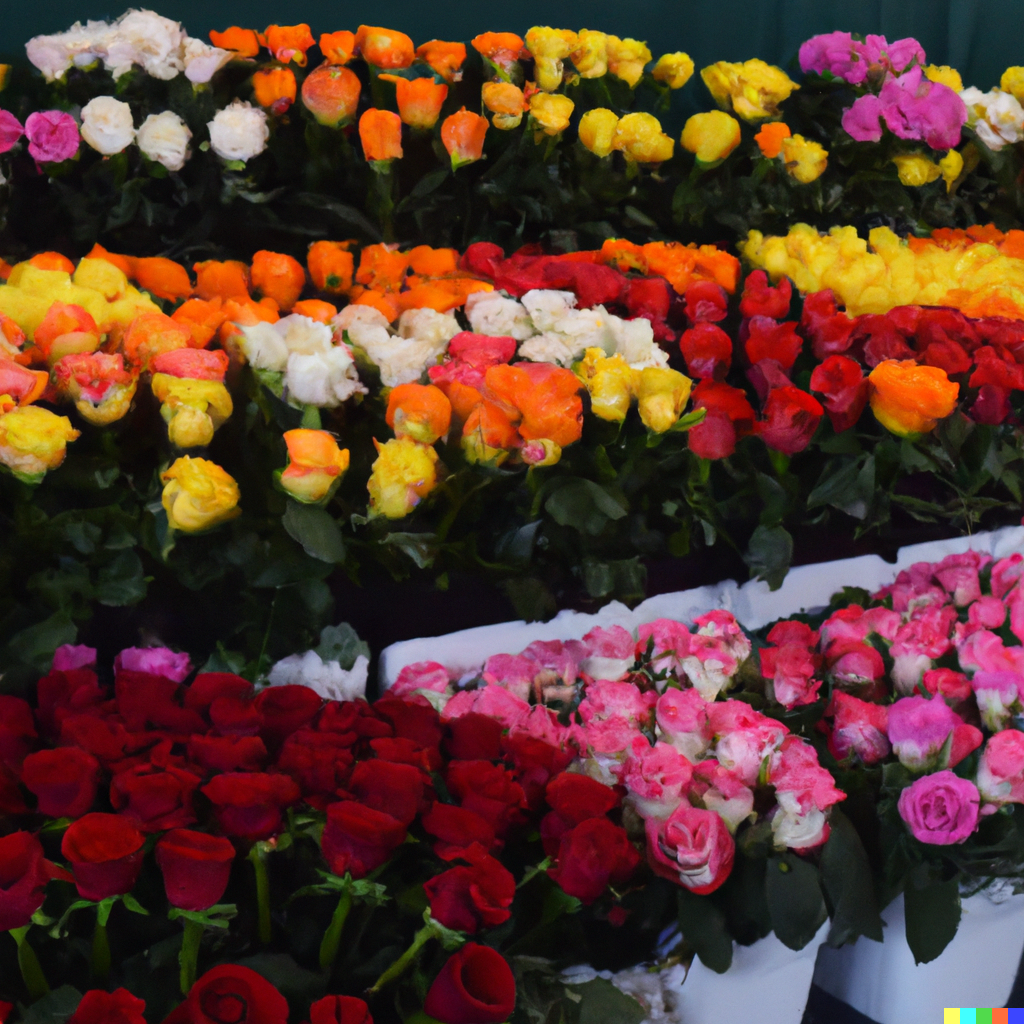} \\
\includegraphics[width=0.50\textwidth]{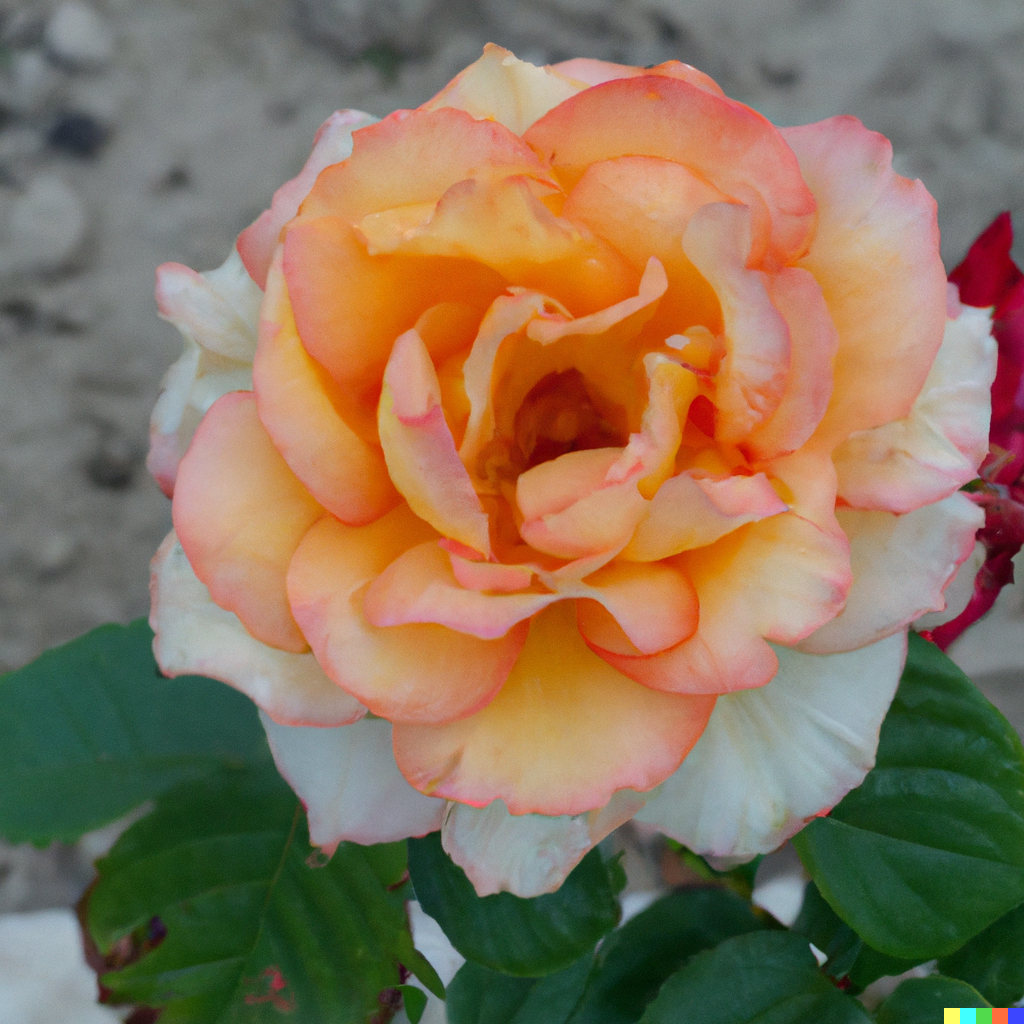} &   
\includegraphics[width=0.50\textwidth]{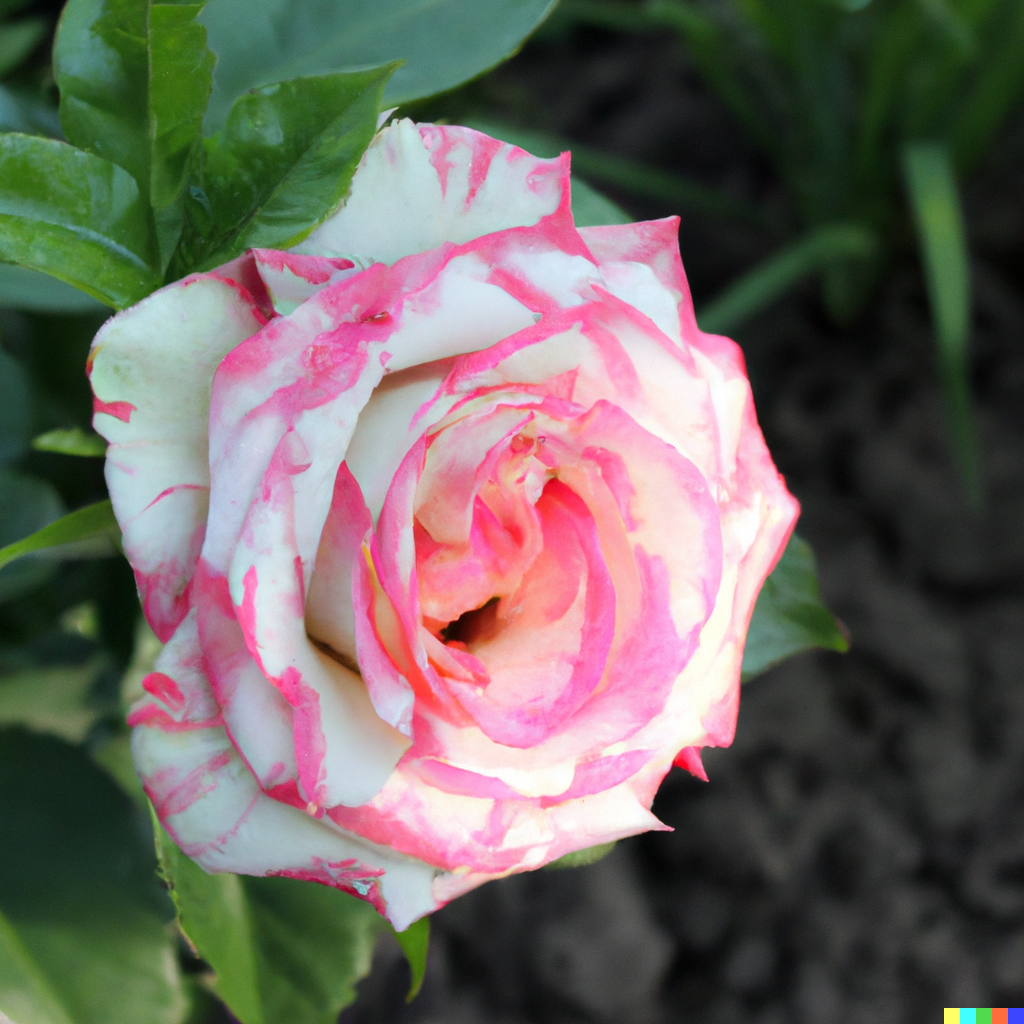} &
\includegraphics[width=0.50\textwidth]{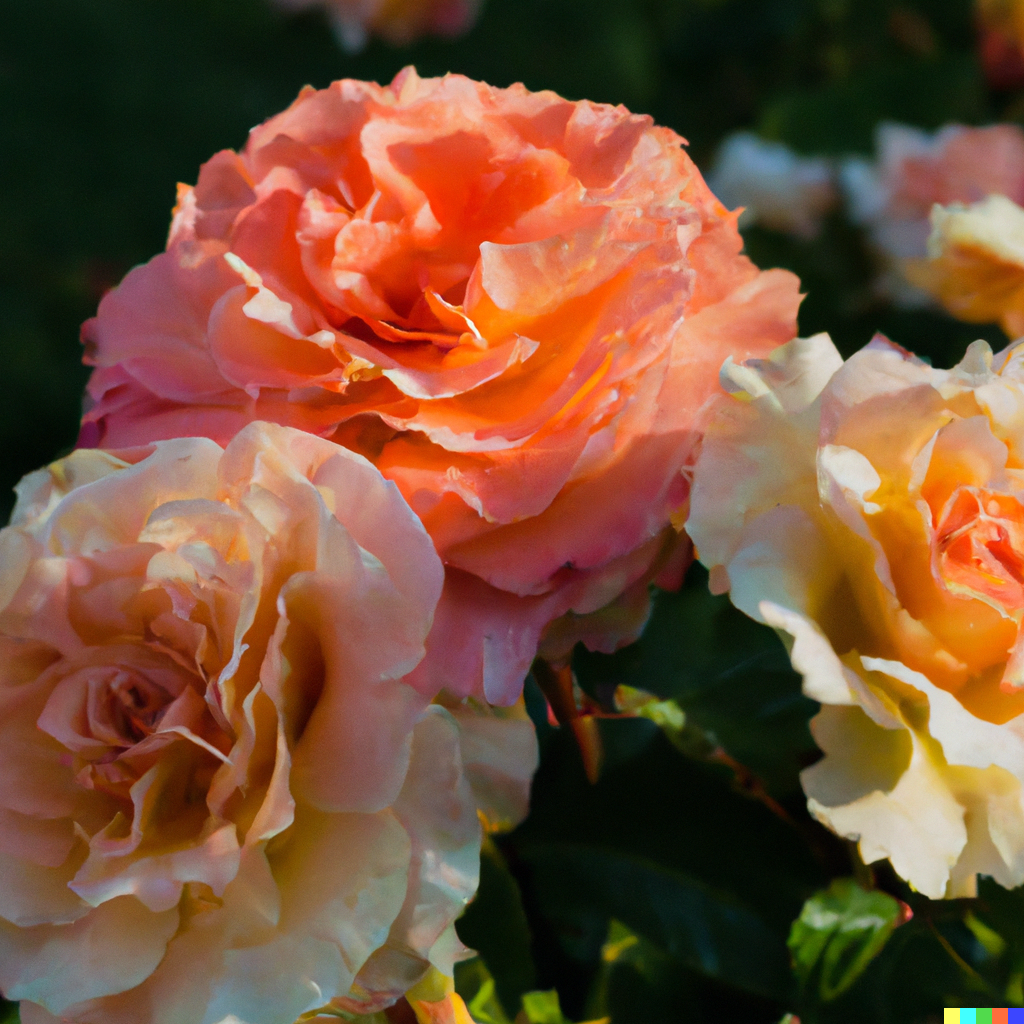} \\  
\end{tabular}}
\caption{Rose Images Generated by Prompting ChatGPT and then CALLE.  The photos of
the first line were generated by the prompt ``generate roses with red, orange, and white colors''. The second line were generated by the prompt ``generate some roses of orange, white, and pink colors with velvety petals in ruffled appearance''. }
\vspace{-.1in}
\label{KGGAN-CALLE}
\end{figure}

A better semantic embedding representation that encodes richer textual 
information about a flower category can be performed by prompting 
a large pre-trained model. The Oxford 
dataset, on which we conducted our experiments, is a tiny knowledge based,
compared to GPT3.  Using GPT3 as our knowledge base, 
we first prompted it for knowledge about roses, and then used the acquired
knowledge to prompt DALLE to generate images.  Figure~\ref{fig:chatgpt-rose} shows
the prompt to query ChatGPT about the colors and textures of roses.  
Once after the color and texture information were obtained, we issued two separate
prompts to DALLE to produce ``roses with red, orange, and white colors'', and 
``roses of orange, white, and pink colors with velvety petals in ruffled appearance''.
The first row of Figure~\ref{KGGAN-CALLE} shows three images of roses with the specified 
colors.  The second row of the figure
shows three rose images with the specified texture specifications.
The ChatGPT and DALLE pipeline can reliably generate a variety of realistic rose
images based on the knowledge acquired from the pre-trained model. When compared
with the flowers generated from a much smaller knowledge-base learned from
the Oxford dataset presented in Figure~\ref{fig:flower_results}, acquiring specifications
from a much larger pre-trained model via ChatGPT clearly 
generates much higher quality rose images.

\section{Concluding Remarks}
\label{sec-conc}

Deep learning has achieved great success thanks to three key factors: the large volume of training data, the ability of advanced models to learn representations from this data, and the scale of computation. In this article, we emphasized the importance of a data-centric approach to learning effective data representations, especially in cases where the available training data is limited.

It is widely accepted in the research community that the more data available, both in terms of volume and variety, the better the performance of representation learning and classification. However, collecting high-quality annotated data in the healthcare domain can be difficult. To address this issue, we discussed four methods for generating and aggregating training data: data augmentation, transfer learning, federated learning, and GANs (generative adversarial networks). While these approaches can be effective, we may still face the challenge of lacking diversity or coverage to represent the full population distribution. To address this problem, we presented knowledge-guided GANs (KG-GANs) as a potential solution. While the idea of KG-GANs was well received by reviewers at top AI conferences, one critique was that knowledge integration must be tailored to the specific task at hand. We believe that recent large pre-trained language models have the potential to serve as a general, task-agnostic knowledge base to support knowledge acquisition. In this article, we demonstrated that the ChatGPT-DALLE pipeline can first acquire precise descriptions of a concept and then use this new ``insight'' to generate realistic images beyond the original, restrictive distribution represented by a small training dataset.

It is currently believed within the artificial intelligence (AI) community that a pre-trained model, trained on all available documents in the world, can serve as a highly accurate and reliable source of knowledge for various tasks. This pre-trained model can then be fine-tuned with a small amount of additional data for a specific task, or prompted step-by-step (e.g., \cite{gao2021prompting,SocraticIEEECCWC2023}) to achieve state-of-the-art performance on that task. These techniques allow for the efficient and effective use of large, pre-trained language models in a variety of applications.

There are several promising directions for future research in the field of artificial intelligence, based on the observations and discussions presented throughout this article. These include:

\begin{itemize}
    \item Improving the interpretability of deep learning models, so that their decision-making processes are more transparent and easier to understand. This is an important consideration for fields such as healthcare, where the consequences of incorrect predictions can be severe.
    \item Combining domain knowledge with existing training data to generate more diverse and representative training data, in order to better cover a wide range of semantic concepts. This could be particularly useful in fields where annotated data is scarce or difficult to collect.
    \item Utilizing large pre-trained models as a knowledge base for robust knowledge acquisition is an important direction for future research. These models, which have been trained on vast amounts of data, can serve as a valuable resource for acquiring knowledge that is generalizable across a wide range of tasks. By leveraging and improving these models, we can increase the reliability and efficiency of knowledge acquisition, which can then be used to guide the generation of more diverse and representative training data for deep learning models. This, in turn, can improve the performance of these models in various applications.
    \item Developing effective techniques for prompting and guiding the acquisition of knowledge, such as using a chain-of-thought or dialogue approach. These methods can help to ensure that the knowledge acquired is precise and relevant to the task at hand.
\end{itemize}

In my opinion, each of these research directions has the potential to significantly advance the field of artificial intelligence and improve the usefulness and reliability of deep learning models in healthcare and a variety of other applications. By focusing on improving the interpretability of deep learning models, incorporating domain knowledge into training data, leveraging large pre-trained models as a knowledge base, and developing effective knowledge prompting techniques, we can make significant progress in enhancing the performance and trustworthiness of these models in healthcare and other fields.

\section*{Acknowledgment}

This article presents the work performed by the DeepQ team between 2014 and 2017 for the Tricorder XPRIZE award \cite{10.1145/3132635.3132637,peng2018refuel}, as well as our research on consciousness modeling at Stanford University since 2020. The relevant papers published by our team~\cite{chang2019kggan,chang2011foundationsch2,KGGAN8695311,2017BookChap1,Transfer-EMBC2015} have been cited throughout the article. We would like to acknowledge the following colleagues for their contributions, listed in alphabetical order: Che-Han Chang, Fu-Chieh Chang, Chun-Nan Chou, Chuen-Kai Shie, and Kai-Fu Tang.

\bibliographystyle{plainnat}
\bibliography{ref,MIPR_new}

\end{document}